\documentclass[11pt]{article}

\usepackage{amsmath,amssymb,amsthm}
\usepackage{bbm}
\usepackage{mathtools}
\usepackage{graphicx}
\usepackage{booktabs}
\usepackage{microtype}
\usepackage{enumitem}
\usepackage{subcaption}
\usepackage{changepage}
\usepackage{setspace}

\usepackage{algorithm}
\usepackage{algorithmic}
\renewcommand{\algorithmicinput}{\textbf{Input:}}
\renewcommand{\algorithmicoutput}{\textbf{Output:}}

\usepackage{tikz}
\usetikzlibrary{arrows.meta,positioning}

\usepackage{arxiv}
\hypersetup{
  pdftitle={Fitted Q-Evaluation without Bellman Completeness via Occupancy Weighting},
  pdfauthor={Lars van der Laan and Nathan Kallus}
}

\DeclareMathOperator*{\argmin}{arg\,min}
\newcommand{\E}{\mathbb{E}}
\newcommand{\ALGCOMMENT}[1]{\item[]\texttt{\#~#1}}

\IfFileExists{generated/fqe/cartpole_fore_fqe_results.tex}{%
\newcommand{\CartPoleN}{50}
\newcommand{\CartPoleUnweightedMAE}{6.000}
\newcommand{\CartPoleFOREMAE}{5.148}
\newcommand{\CartPoleRelativeReduction}{14.2\%}
\newcommand{\CartPoleGainCILower}{0.611}
\newcommand{\CartPoleGainCIUpper}{1.087}
\newcommand{\CartPoleFOREWins}{44}
\newcommand{\CartPoleMedianESS}{0.759}

}{%
  \newcommand{\CartPoleN}{50}
  \newcommand{\CartPoleUnweightedMAE}{\textsc{pending}}
  \newcommand{\CartPoleFOREMAE}{\textsc{pending}}
  \newcommand{\CartPoleRelativeReduction}{\textsc{pending}}
  \newcommand{\CartPoleGainCILower}{\textsc{pending}}
  \newcommand{\CartPoleGainCIUpper}{\textsc{pending}}
  \newcommand{\CartPoleFOREWins}{\textsc{pending}}
  \newcommand{\CartPoleMedianESS}{\textsc{pending}}
  
}

\theoremstyle{plain}
\newtheorem{theorem}{Theorem}
\newtheorem{corollary}{Corollary}
\newtheorem{lemma}{Lemma}
\newtheorem{proposition}{Proposition}
\theoremstyle{definition}

\theoremstyle{remark}
\newtheorem{remark}{Remark}

\title{Fitted \(Q\)-Evaluation without Bellman Completeness\\
via Occupancy Weighting}
\author{%
  Lars van der Laan \\
  Department of Management Science and Engineering, Stanford University \\
  \texttt{vdlaan@stanford.edu} \\
  \And
  Nathan Kallus \\
  Netflix and Cornell University
}
\begin{document}

\maketitle

\begin{abstract}
Fitted \(Q\)-evaluation (FQE) is a standard regression-based method for
off-policy evaluation, but under distribution shift, value-function
realizability alone does not ensure convergence, and existing analyses often
require Bellman completeness. We trace this instability to a geometric
mismatch: standard FQE projects Bellman targets in the norm induced by the
offline distribution, which need not preserve Bellman contraction. We
therefore study \emph{occupancy-weighted FQE}, which changes only the
regression weights. Weighting by a target-policy discounted occupancy ratio
aligns the projection norm with the target-policy dynamics and restores
contraction of the population projected Bellman operator. We derive
finite-sample guarantees with estimated occupancy ratios and function-class
misspecification, separating finite-iteration, statistical, approximation, and
ratio-estimation errors. Exact occupancy weighting removes the need for Bellman
completeness; with estimated weights, approximate completeness and
value-function realizability reduce sensitivity to ratio-estimation error,
with exact realizability yielding higher-order dependence. Combining
occupancy-weighted FQE with fitted occupancy-ratio evaluation gives an
end-to-end guarantee governed by the complexities and direct approximation
errors of the value-function and occupancy-ratio classes. Under coverage,
joint realizability of these two classes suffices for consistent estimation
without Bellman or critic-side completeness. Controlled experiments illustrate
the projection-norm mechanism and the finite-sample tradeoff between
contraction and coverage. 
\end{abstract}

\section{Introduction}

Fitted $Q$-evaluation (FQE) is a standard regression-based method for policy
evaluation in reinforcement learning
\citep{tsitsiklis1996analysis,gordon1995stable,lagoudakis2003least,
ernst2005tree,munos2005error,munos2008finite,farahmand2010error,
bertsekas2011approximate}. FQE approximates Bellman fixed-point iteration by
repeatedly forming one-step targets from the current value-function estimate
and regressing them onto a function class. It is therefore simple to implement
and readily accommodates off-the-shelf supervised learning methods
\citep{voloshin2019empirical,fujimoto2019off,le2019batch,agarwal2021deep}.
Under off-policy distribution shift, however, value-function realizability
alone does not ensure convergence, and FQE can fail even with linear function
approximation
\citep{baird1995residual,gordon1995stable,tsitsiklis1996analysis}.

This instability arises from a mismatch between the geometry of regression and
that of Bellman contraction. At the population level, FQE iterates a projected
Bellman operator: each Bellman update is projected onto the fitted function
class in the $L^2$ norm induced by the offline sampling distribution. The
Bellman operator, by contrast, is contractive in the supremum norm and in
target-policy occupancy norms, but not generally in the offline-sampling norm
\citep{tsitsiklis1996analysis,patterson2022generalized}. The resulting
projected Bellman operator therefore need not be contractive, and the classical
convergence guarantees for exact Bellman iteration no longer apply.

Contractivity, and hence geometric convergence, is commonly obtained through
\textit{Bellman completeness}, which requires the Bellman operator to map the
fitted value-function class into itself. Projection is then inactive on
Bellman images, so FQE inherits the contraction of exact Bellman iteration.
Related analyses replace exact completeness with direct stability assumptions
on the projected Bellman operator or with approximate completeness conditions
that bound the \emph{inherent Bellman error}, namely, how far Bellman images of
functions in the fitted class lie outside that class
\citep{antos2007fitted,munos2008finite,amortila2020variant,
wang2021exponential,wang2021statistical,chang2022learning}.

Bellman completeness is restrictive under function approximation and behaves
unlike a standard approximation condition. In particular, it is non-monotone
in the fitted class: enlarging the class can reduce ordinary approximation
error while worsening completeness, because the Bellman images of newly added
functions must also be represented
\citep{zhan2022offline,zanette2023realizability}. More fundamentally, the need
for structure beyond value-function realizability and coverage is not merely a
proof artifact. \citet{chen2019information} highlighted the gap between
realizability and Bellman completeness, while \citet{foster2021offline} showed
that standard concentrability and value-function realizability alone are
insufficient for sample-efficient offline policy evaluation and learning.
Additional structure is therefore unavoidable in the worst case, although it
need not take the form of Bellman completeness.

Minimax and adversarial \(Q\)-learning provide an alternative, estimating
value functions through saddle-point formulations of Bellman moment conditions
\citep{dai2018sbeed,uehara2020minimax,feng2020accountable,xie2020q,
xie2021bellman,uehara2023offline,chen2022well,zanette2022bellman}.
These methods avoid requiring the value-function class to be Bellman complete
but shift structural requirements to an auxiliary critic or test-function
class. Their guarantees typically require the critic class to be sufficiently
rich to detect or control Bellman errors generated by candidate value
functions, through critic realizability, completeness, or related richness
conditions
\citep{feng2019kernel,jiang2020minimax,uehara2021finite,
zanette2022bellman,zhan2022offline}.

FQE nevertheless remains widely used for offline policy evaluation because it
is simple to implement and readily accommodates flexible supervised-learning
methods \citep{voloshin2019empirical,tang2021model}. Rather than replacing FQE
with a different estimation framework, we ask whether its instability can be
corrected through a minimal modification. Our approach builds on work that
aligns value-function updates with target-policy visitation distributions to
stabilize off-policy temporal-difference learning
\citep{hallak2017consistent,geladaBellemare2019CovariateShift,
sinha2022experience}. We adapt this principle to offline FQE by weighting each
Bellman regression by the density ratio of the target-policy occupancy to the
offline distribution. In a
\(\gamma\)-discounted MDP, we use the \(\beta\)-discounted target-policy
occupancy. For every
\(\beta\in(\gamma^2,1)\), this makes the population projected Bellman operator
a \(\gamma/\sqrt{\beta}\)-contraction in the corresponding \(L^2\) norm. The
method therefore corrects FQE's projection geometry while preserving its
supervised-learning structure and requires neither an auxiliary critic class
nor joint value--critic saddle-point optimization.

We call the resulting procedure \emph{occupancy-weighted FQE}. Our main
contribution is a finite-sample analysis under function-class misspecification
and estimated weights that separates finite-iteration, statistical,
approximation, and ratio-estimation errors. The analysis shows that
ratio-estimation error matters through its interaction with how well Bellman
updates can be represented by the fitted function class. Exact occupancy
weighting removes this interaction. With estimated weights, approximate
Bellman completeness controls it uniformly over the fitted class, whereas
approximate value-function realizability provides a sharper local control near
the projected fixed point; under exact realizability, the dependence on
ratio-estimation error becomes higher order. Thus, the guarantees interpolate
between exact distribution correction and structural control through
completeness or realizability.

\begin{enumerate}
\item \textbf{Occupancy-weighted FQE.}
We show that target-policy occupancy weighting restores contraction of fitted
policy evaluation in an appropriate \(L^2\) norm and establish finite-sample
guarantees with estimated weights and function-class misspecification, without
requiring Bellman completeness.

\item \textbf{Complementary roles of completeness and realizability.}
We show that weight-estimation error interacts with the Bellman projection
residual. Approximate Bellman completeness and approximate value-function
realizability attenuate this interaction through uniform and local control,
respectively; exact completeness removes the weight-error contribution,
whereas exact realizability makes it higher order.

\item \textbf{End-to-end estimation.}
Combining occupancy-weighted FQE with fitted occupancy-ratio evaluation
(FORE) yields guarantees governed by the complexities and direct approximation
errors of the value-function and occupancy-ratio classes. Under coverage,
joint realizability leaves only statistical and finite-iteration errors,
without Bellman completeness, critic-side completeness, or auxiliary critic
classes.
\end{enumerate}

\subsection{Related Work}

\paragraph{Fitted Bellman iteration.}
Classical analyses of fitted value iteration and FQE establish stability of the
projected Bellman iteration or bound the propagation of one-step approximation
and estimation errors
\citep{gordon1995stable,tsitsiklis1996analysis,munos2005error,
munos2008finite,farahmand2010error,bertsekas2011approximate,
scherrer2014approximate}. Under function approximation, these guarantees
typically impose additional structure linking the Bellman operator, fitted
class, and sampling distribution, including Bellman completeness, small
inherent Bellman error, uniform coverage or concentrability conditions, or
structured representations
\citep{antos2007fitted,chen2019information,amortila2020variant,
wang2021statistical,chang2022learning,van2006performance,xie2021batch}.
Occupancy weighting instead aligns the regression geometry with a
target-occupancy \(L^2\) norm, yielding guarantees in terms of approximation
of \(Q^\pi\) itself rather than uniform approximation of Bellman images.

This distinction persists under linear function approximation: projection in
an arbitrary off-policy norm need not preserve contraction, and the projected
fixed point can be arbitrarily inaccurate despite small value-function
approximation error \citep{yu2010error,kolter2011fixed}. Our results are
therefore most relevant beyond tabular, linear-MDP, and Bellman-complete
settings, where fitted iteration is already controlled by stronger structural
assumptions
\citep{melo2007convergence,yang2020reinforcement,jin2023provably}.
 
\paragraph{Weighted temporal-difference and projected Bellman methods.}
Our approach builds on work that stabilizes off-policy value learning by
changing the effective update distribution. Under linear function
approximation, off-policy instability motivated emphatic methods based on
follow-on weights
\citep{mahmood2015emphatic,sutton2016emphatic,yu2015convergence,
yu2018generalized} and distribution-correction methods based on stationary
density ratios
\citep{hallak2017consistent,geladaBellemare2019CovariateShift,
sinha2022experience}. Related work studies alternative projection norms and
generalized projected Bellman objectives
\citep{yu2012weighted,van2006performance,patterson2022generalized}, including
nonlinear methods based on auxiliary Bellman-error functions and
gradient-correction or saddle-point updates.

Occupancy-weighted FQE applies the same geometric principle to a different
recursion: successive supervised regressions against frozen Bellman targets.
Outside linear function approximation, this fitted iteration is distinct from
TD-style stochastic approximation and requires separate analysis. We use the
density ratio of the target-policy occupancy to the offline distribution only
to set the regression geometry, preserving the standard FQE update otherwise.
Our finite-sample
analysis allows nonlinear function classes, misspecification, and estimated
weights, and isolates how ratio-estimation error interacts with the Bellman
projection residual.

\paragraph{Model-based off-policy evaluation.}
Model-based OPE instead fits a transition model and evaluates the target policy
inside that model. Minimax model learning changes the model-fitting loss to
control target-policy value error under distribution shift and model
misspecification \citep{voloshin2021minimax}. Occupancy-weighted FQE is
model-free: it estimates only the target-to-offline occupancy ratio and retains
the observed one-step transitions in every Bellman regression.

\paragraph{Minimax and adversarial \(Q\)-learning.}
Minimax and adversarial methods replace fitted Bellman iteration with
saddle-point objectives that test Bellman residuals against an auxiliary critic
class
\citep{dai2018sbeed,uehara2020minimax,feng2020accountable,xie2020q,
xie2021bellman,imaizumi2021minimax,jin2021bellman,
uehara2023offline,zanette2022bellman}. These methods can avoid Bellman
completeness of the value-function class, but their guarantees generally
require the critic class to be sufficiently rich to detect the Bellman errors
generated by candidate value functions, for example through conditions of the
form
\(\{\mathcal TQ-Q:Q\in\mathcal Q\}\subseteq c\mathcal G\)
\citep{uehara2021finite}. Thus, the structural requirement shifts from
Bellman completeness of the value class to critic-side completeness or
richness.

Occupancy-weighted FQE instead uses no auxiliary critic class. Combined with
FORE, joint realizability of the \(Q\)-function and occupancy density ratio
suffices for consistent estimation of both functions under coverage. This is
stronger than some minimax OPE guarantees under
joint realizability, which establish consistency of the scalar policy value
without requiring consistent nuisance estimation \citep{uehara2021finite}.
Consistent nuisance recovery also enables semiparametrically efficient
policy-value estimation through a doubly robust estimator under standard
product-rate conditions \citep{kallus2020double}; see
Appendix~\ref{app:discounted-occupancy-fqe}.

\subsection{Preliminaries}

We consider a discounted MDP with general state space \(\mathcal S\), finite
action space \(\mathcal A\), transition kernel \(P(\cdot\mid s,a)\), mean reward
\(r_0(s,a)\), and discount factor \(\gamma\in[0,1)\). A fixed target policy
\(\pi(a\mid s)\) induces the state-action transition operator
\[
(P_\pi Q)(s,a)
:=
\mathbb E\!\left[Q(S',A')\mid S=s,A=a\right],
\qquad
S'\sim P(\cdot\mid s,a),\quad A'\sim\pi(\cdot\mid S').
\]
The Bellman operator for \(\pi\) and its unique fixed point are
\[
\mathcal TQ:=r_0+\gamma P_\pi Q,
\qquad
Q^\pi=\mathcal TQ^\pi.
\]
The fixed point \(Q^\pi\) is the target-policy \(Q\)-function, giving the
expected discounted return after taking action \(a\) in state \(s\) and
following \(\pi\) thereafter.

We observe transitions
\(\mathcal D_n=\{(S_i,A_i,R_i,S_i')\}_{i=1}^n\), where
\(S_i'\sim P(\cdot\mid S_i,A_i)\) and
\(R_i=r_0(S_i,A_i)+\varepsilon_i\), with
\(\mathbb E[\varepsilon_i\mid S_i,A_i]=0\). We assume that
\((S_i,A_i)\) have common marginal distribution \(\nu_b\), while the
transition tuples may be dependent. We refer to \(\nu_b\) as the offline
state-action distribution; under stationary trajectory sampling, it is the
stationary state-action distribution induced by the behavior policy. For any
state-action distribution \(\nu\), define
\[
(\pi f)(s)
:=
\sum_{a\in\mathcal A}\pi(a\mid s)f(s,a),
\qquad
\|f\|_{2,\nu}
:=
\left\{
\int f(s,a)^2\,\nu(ds,da)
\right\}^{1/2}.
\]

 \subsection{Standard FQE and Norm Mismatch}

Classical policy evaluation computes \(Q^\pi\) by iterating the Bellman
operator \(\mathcal T\)
\citep{bellman1952theory,bellman1966dynamic}. Because \(\mathcal T\) is a
\(\gamma\)-contraction in the supremum norm, the iterates converge
geometrically to \(Q^\pi\). With function approximation and offline data, FQE
replaces exact Bellman iteration by fitted regression
\citep{lagoudakis2003least,ernst2005tree,munos2008finite}. Given a function
class \(\mathcal F\), its empirical update is
\[
\widehat{Q}^{(k+1)}
\in
\arg\min_{f\in\mathcal F}
\frac{1}{n}\sum_{i=1}^n
\left\{
R_i+\gamma(\pi\widehat{Q}^{(k)})(S_i')
-f(S_i,A_i)
\right\}^2.
\]
At the population level, FQE iterates the projected Bellman operator
\(\Pi_{\mathcal F}^{\nu_b}\mathcal T\), where
\(\Pi_{\mathcal F}^{\nu_b}\) denotes \(L^2(\nu_b)\) projection onto
\(\mathcal F\). Unlike \(\mathcal T\), this projected operator need not be
contractive.

A common way to recover stability is Bellman completeness,
\(\mathcal T\mathcal F\subseteq\mathcal F\). Then projection is inactive on
Bellman images:
\(\Pi_{\mathcal F}^{\nu_b}\mathcal T f=\mathcal T f\) for every
\(f\in\mathcal F\), so FQE inherits the contraction of exact Bellman
iteration. Bellman completeness is therefore central to many analyses of FQE.
Other analyses instead assume stability of the projected Bellman operator
directly or control the inherent Bellman error
\(\sup_{f\in\mathcal F}
\|\mathcal T f-\Pi_{\mathcal F}^{\nu_b}\mathcal T f\|_{L^2(\nu_b)}\)
\citep{antos2007fitted,munos2008finite,chen2019information,
amortila2020variant,wang2021exponential,
wang2021statistical,chang2022learning}.

These conditions address the same underlying difficulty: regression and
Bellman updates are naturally controlled in different geometries.
Least-squares projection is defined in \(L^2(\nu_b)\) and, for closed convex
\(\mathcal F\), is nonexpansive in that norm. By contrast, the Bellman
operator is contractive in the supremum norm and in suitable target-policy
occupancy norms \citep{tsitsiklis1996analysis}. Off policy, the offline and
target-occupancy geometries may be misaligned, so
\(\Pi_{\mathcal F}^{\nu_b}\mathcal T\) need not inherit Bellman contraction.

\section{Occupancy-Weighted Fitted \(Q\)-Evaluation}

\subsection{Population Geometry and Contraction}
\label{sec::stable}

We resolve this mismatch by projecting Bellman updates in a target-policy
occupancy norm. Let \(\nu_b^S\) denote the state marginal of \(\nu_b\), and
define the target-policy restart distribution
\[
\xi_{\pi,b}(ds,da)
:=
\nu_b^S(ds)\pi(da\mid s).
\]
For \(\beta\in(0,1)\), define the corresponding discounted occupancy
\begin{equation}
\label{eq:offline-restarted-occupancy}
d_\beta
:=
(1-\beta)\sum_{t=0}^{\infty}\beta^t\xi_{\pi,b}P_\pi^t,
\qquad
d_\beta
=
(1-\beta)\xi_{\pi,b}+\beta d_\beta P_\pi.
\end{equation}
Define the projected Bellman operator
\begin{equation}
\label{eqn:projbell}
\mathcal T_{\mathcal F,\beta}
:=
\Pi_{\mathcal F,\beta}\mathcal T,
\qquad
\Pi_{\mathcal F,\beta}g
:=
\argmin_{f\in\mathcal F}
\|f-g\|_{2,d_\beta}.
\end{equation}
The corresponding population fitted iteration is
\begin{equation}
\label{eqn:projbell2}
Q_{\mathcal F,\beta}^{(k)}
=
\mathcal T_{\mathcal F,\beta}Q_{\mathcal F,\beta}^{(k-1)}.
\end{equation}

\begin{enumerate}[label=\textbf{C\arabic*}, ref={C\arabic*},
leftmargin=1.5em, series=cond]
\item \label{cond::stationary}
\textbf{Occupancy discount.}
The discount satisfies \(\beta\in(\gamma^2,1)\), and
\(r_0\in L^2(d_\beta)\).

\item \label{cond::convex}
\textbf{Closed convex class.}
The class \(\mathcal F\) is closed and convex in \(L^2(d_\beta)\).
\end{enumerate}

The restriction \(\beta>\gamma^2\) ensures that the contraction factor below
is strictly less than one. In particular, \(\beta=\gamma\) always satisfies
this condition and yields contraction factor \(\sqrt{\gamma}\).
Condition~\ref{cond::convex} ensures that the metric projection is uniquely
defined and nonexpansive \citep{bauschke2011convex}.

\begin{lemma}[Occupancy-weighted projected Bellman contraction]
\label{lem:bellman-contraction}
Under Conditions~\ref{cond::stationary}--\ref{cond::convex}, for all
\(Q,Q'\in\mathcal F\),
\[
\|\mathcal T_{\mathcal F,\beta}Q-\mathcal T_{\mathcal F,\beta}Q'\|_{2,d_\beta}
\le
\bar\gamma_\beta\|Q-Q'\|_{2,d_\beta},
\qquad
\bar\gamma_\beta:=\frac{\gamma}{\sqrt{\beta}}<1.
\]
\end{lemma}

By Banach's fixed-point theorem, \(\mathcal T_{\mathcal F,\beta}\) has a
unique fixed point \(Q^\pi_{\mathcal F,\beta}\in\mathcal F\), and the
population recursion converges geometrically. Moreover,
\[
\begin{aligned}
\|Q_{\mathcal F,\beta}^{(K)}-Q^\pi_{\mathcal F,\beta}\|_{2,d_\beta}
&\le
\bar\gamma_\beta^K
\|Q_{\mathcal F,\beta}^{(0)}-Q^\pi_{\mathcal F,\beta}\|_{2,d_\beta},\\
\|Q^\pi_{\mathcal F,\beta}-Q^\pi\|_{2,d_\beta}
&\le
\frac{1}{1-\bar\gamma_\beta}
\inf_{f\in\mathcal F}\|f-Q^\pi\|_{2,d_\beta}.
\end{aligned}
\]
Thus \(Q^\pi_{\mathcal F,\beta}=Q^\pi\) under realizability; under
misspecification, the fixed-point error is controlled by the best
\(L^2(d_\beta)\) approximation of \(Q^\pi\) in \(\mathcal F\).

If \(d_\beta\ll\nu_b\), let
\(w_\beta:=\mathrm d d_\beta/\mathrm d\nu_b\). The same projection can then
be implemented as weighted regression under the offline distribution:
\[
\mathcal T_{\mathcal F,\beta}Q
=
\argmin_{f\in\mathcal F}
\E_{\nu_b}\!\left[
w_\beta(S,A)
\left\{
R+\gamma(\pi Q)(S')-f(S,A)
\right\}^2
\right].
\]
Thus occupancy weighting changes the regression norm while preserving the
standard FQE regression objective.

\textbf{Choosing the occupancy discount.}
Larger \(\beta\) improves the contraction modulus
\(\bar\gamma_\beta=\gamma/\sqrt{\beta}\), approaching \(\gamma\) as
\(\beta\uparrow1\), but may also increase weight variability as the effective
horizon grows. The choice \(\beta=\gamma\) gives modulus
\(\sqrt{\gamma}\) and is a natural default because \(\gamma\)-discounted
occupancies also arise in policy-value estimation, including doubly robust
estimators \citep{kallus2020double}. When \(d_\beta\) converges to the
target-policy stationary distribution as \(\beta\uparrow1\), the classical
stationary-distribution contraction with modulus \(\gamma\) is recovered
\citep{tsitsiklis1996analysis,hallak2017consistent,van2006performance}.
We use the offline state marginal as the restart distribution because it
yields a direct coverage guarantee through action overlap, although the
contraction argument applies to any restart state distribution.

\textbf{Convexity.}
Condition~\ref{cond::convex} covers closed convex subsets of linear, RKHS, and
classical smoothness classes \citep{van1996weak}, as well as convex
neural-network classes \citep{bach2017breaking}. Standard finite-width neural
network classes are nonconvex and therefore fall outside the present
projection argument. Nevertheless, neural networks of increasing complexity
can approximate functions in the convex smoothness classes covered by the
theory \citep{suzuki2019adaptivity}. A direct extension to nonconvex
neural-network classes would additionally require control of optimization
error.

\subsection{Empirical Occupancy-Weighted FQE}
\label{sec:sw-fqe-algorithm}

Replacing the population objective and oracle ratio with empirical
counterparts yields occupancy-weighted FQE. Given an estimate
\(\widehat w_\beta\) of \(w_\beta\) and an initialization
\(\widehat Q^{(0)}\in\mathcal F\), iterate
\begin{equation}
\label{eq:sw-fqe-update}
\widehat Q^{(k+1)}
\in
\argmin_{f\in\mathcal F}
\frac{1}{n}\sum_{i=1}^{n}
\widehat w_\beta(S_i,A_i)
\left\{
R_i+\gamma(\pi\widehat Q^{(k)})(S_i')
-f(S_i,A_i)
\right\}^2.
\end{equation}
With oracle weights, the population minimizer is
\(\mathcal T_{\mathcal F,\beta}\widehat Q^{(k)}\). Relative to standard FQE, the
Bellman targets are unchanged; only the regression loss is reweighted. Each
iteration therefore remains a weighted supervised regression, with the density
ratio estimated once and reused across iterations.

The regressions may be fit using any method that supports observation weights.
For neural networks, for example, one may take one or more minibatch gradient
steps with fixed Bellman targets before refreshing them, as in practical fitted
value iteration \citep{mnih2013playing}.

\subsection{Estimation of the Occupancy Density Ratio}
\label{sec:ratio-estimation}

Occupancy-weighted FQE requires an estimate of the density ratio
\(w_\beta=\mathrm d d_\beta/\mathrm d\nu_b\). Whereas occupancy ratios are
typically used to reweight rewards or construct policy-value estimators
\citep{jiang2016doubly,thomas2016data,kallus2020double,kallus2022efficiently,
van2025automaticDRL}, here \(w_\beta\) determines the projection norm in each
Bellman regression.

The ratio satisfies the occupancy-balance equation
\citep{uehara2021finite}: for every bounded measurable \(g\),
\begin{equation}
\label{eq:weak-occupancy-balance}
\mathbb E_{\nu_b}\!\left[w_\beta(S,A)g(S,A)\right]
=
(1-\beta)\mathbb E_{\xi_{\pi,b}}[g(S,A)]
+
\beta\mathbb E_{\nu_b}\!\left[w_\beta(S,A)g(S',A')\right].
\end{equation}
Here \(S'\sim P(\cdot\mid S,A)\),
\(A'\sim\pi(\cdot\mid S')\), and
\(\mathbb E_{\nu_b}[w_\beta(S,A)]=1\).
Existing estimators exploit these restrictions through stationary correction,
primal--dual or minimax objectives, and projected balancing methods
\citep{glynn1998estimation,liu2018breaking,nachum2019dualdice,
zhang2020gendice,uehara2020minimax,uehara2021finite,
wang2023projected,amortila2024harnessing}.

Our FQE analysis is agnostic to the ratio estimator and requires only control
of its estimation error. Minimax and DICE-style estimators typically introduce
an auxiliary critic class, with guarantees depending on its richness relative
to the ratio class \citep{uehara2021finite}. For an end-to-end procedure,
Algorithm~\ref{alg:fore-weighted-fqe} uses fitted occupancy-ratio evaluation
(FORE) \citep{van2026fitted}, whose population update contracts in generalized
KL under the offline distribution. Under coverage, its approximation error
depends only on direct approximation of the true occupancy ratio by the ratio
class and therefore vanishes under ratio realizability, without an auxiliary
critic class or critic-side completeness requirement.

For theoretical convenience, we use independent samples
\(\mathcal D_{n_w}^w=\{(S_i,A_i,S_i')\}_{i=1}^{n_w}\) for ratio estimation
and \(\mathcal D_n^Q\) for FQE. For each \(i\), set
\(X_i=(S_i,A_i)\), draw
\(A_i^0\sim\pi(\cdot\mid S_i)\) and
\(A_i^+\sim\pi(\cdot\mid S_i')\), and define
\(X_i^0=(S_i,A_i^0)\) and \(X_i^+=(S_i',A_i^+)\).
In practice, the ratio may instead be estimated on the same sample or
cross-fitted to reuse data.

\begingroup
\setlength{\intextsep}{6pt}
\begin{algorithm}[H]
\caption{\textbf{FORE-Weighted Fitted \(Q\)-Evaluation}}
\label{alg:fore-weighted-fqe}
\begin{algorithmic}[1]
\small
\INPUT Samples \(\mathcal D_{n_w}^w,\mathcal D_n^Q\), policy \(\pi\),
classes \(\mathcal G_w,\mathcal F\), iterations \(K_w,K\), discounts
\(\gamma,\beta\)
\STATE Initialize \(\widehat w^{(0)}\equiv1\)
\FOR{\(\ell=0,\ldots,K_w-1\)}
\STATE \(\displaystyle
\widehat g^{(\ell+1)}
\in
\argmin_{g\in\mathcal G_w}
\left[
\log\left\{
\frac{1}{n_w}\sum_{i=1}^{n_w}e^{g(X_i)}
\right\}
-(1-\beta)\frac1{n_w}\sum_{i=1}^{n_w}g(X_i^0)
-\beta
\frac{
\sum_{i=1}^{n_w}
\widehat w^{(\ell)}(X_i)g(X_i^+)
}{
\sum_{i=1}^{n_w}
\widehat w^{(\ell)}(X_i)
}
\right]
\)
\STATE \(\displaystyle
\widehat w^{(\ell+1)}(x)
\gets
\frac{
e^{\widehat g^{(\ell+1)}(x)}
}{
n_w^{-1}\sum_{i=1}^{n_w}
e^{\widehat g^{(\ell+1)}(X_i)}
}
\)
\ENDFOR
\STATE Set \(\widehat w_\beta\gets\widehat w^{(K_w)}\), initialize
\(\widehat Q^{(0)}\in\mathcal F\), and apply
\eqref{eq:sw-fqe-update} for \(k=0,\ldots,K-1\) on \(\mathcal D_n^Q\)
\OUTPUT \(\widehat w_\beta,\widehat Q^{(K)}\)
\end{algorithmic}
\end{algorithm}
\endgroup

\section{Theoretical Guarantees}
\label{sec:theorymain}

We analyze the exact empirical-risk-minimization version of FORE-weighted FQE
in Algorithm~\ref{alg:fore-weighted-fqe}. Each empirical update approximates
the population projected Bellman map
\(\mathcal T_{\mathcal F,\beta}\), whose fixed point is
\(Q^\pi_{\mathcal F,\beta}\). The finite-sample analysis therefore reduces to
controlling the one-step fitted-regression error, including the additional
error induced by estimating the occupancy weights. We first bound
\[
\left\|
\widehat Q^{(k)}
-
\mathcal T_{\mathcal F,\beta}\widehat Q^{(k-1)}
\right\|_{2,d_\beta},
\]
and then propagate this error through the population contraction.

\subsection{One-Step Regression Error under Approximate Weighting}
\label{sec:theory:reg}

The one-step analysis separates statistical regression error from
weight-estimation error, with the latter coupled to the Bellman projection
residual of the fitted class. We impose the following regularity conditions.
\begin{enumerate}[label=\textbf{C\arabic*}, ref={C\arabic*},
leftmargin=1.5em, resume=cond]
\item \label{cond::bounded}
\textbf{Boundedness.}
There exists \(M<\infty\) such that \(|R|\le M\) and
\(\sup_{f\in\mathcal F}\|f\|_{\infty}\le M\).

\item \label{cond::independent-transitions}
\textbf{Independent transitions.}
The transition tuples
\(\{(S_i,A_i,R_i,S_i')\}_{i=1}^n\) are independent, with
\((S_i,A_i)\sim\nu_b\).

\item \label{cond::split}
\textbf{Sample splitting.}
The weight estimator \(\widehat w_\beta\) is computed from data independent of
\(\mathcal D_n\).
\end{enumerate}
Conditions~\ref{cond::bounded}--\ref{cond::split} streamline the
finite-sample analysis. Following \citet{chen2019information},
Condition~\ref{cond::independent-transitions} assumes independent transitions;
extensions to mixing trajectories follow by replacing i.i.d. concentration
bounds with their mixing counterparts \citep{jiangSunFan2018Bernstein}.
Condition~\ref{cond::split} separates weight estimation from Bellman
regression, yielding fast ERM rates without controlling the complexity of the
weight estimator \citep{foster2023orthogonal}. The analysis extends directly
to cross-fitting, allowing all observations to contribute to both stages.
Without sample splitting, the rate may additionally depend on the complexity
or stability of the weight estimator \citep{van2026researcher}.

We next impose coverage and estimated-weight regularity.

\begin{enumerate}[label=\textbf{C\arabic*}, ref={C\arabic*},
leftmargin=1.5em, resume=cond]
\item \label{cond::weight-coverage}
\textbf{Coverage and weight bounds.}
There
exists \(1\le\kappa_{\mathrm{cov}}<\infty\) such that, \(\nu_b\)-almost surely,
\[
w_\beta(S,A)\le\kappa_{\mathrm{cov}},
\qquad
0\le\widehat w_\beta(S,A)\le\kappa_{\mathrm{cov}}.
\]

\item \label{cond::weight-stability}
\textbf{Restricted norm comparability.}
Let \(\lambda_{\mathcal F}>0\) be any constant such that, for every
\(f,f'\in\mathcal F\),
\[
\mathbb E_{\nu_b}\!\left[\widehat w_\beta(f-f')^2\right]
\ge
\lambda_{\mathcal F}
\mathbb E_{\nu_b}\!\left[w_\beta(f-f')^2\right]
=
\lambda_{\mathcal F}
\|f-f'\|_{2,d_\beta}^2.
\]
\end{enumerate}
Condition~\ref{cond::weight-coverage} imposes the standard
overlap or concentrability bound on \(w_\beta\)
\citep{xie2022role,uehara2020minimax}, while boundedness of
\(\widehat w_\beta\) can be enforced by clipping or constrained estimation.
Condition~\ref{cond::weight-stability} prevents the norm induced by the
estimated weights from becoming degenerate over directions in
\(\mathcal F-\mathcal F\). A simple sufficient condition is that the estimated
weights are uniformly bounded away from zero: if
\(\widehat w_\beta\ge \underline w>0\), then
Condition~\ref{cond::weight-stability} holds with
\(\lambda_{\mathcal F}=\underline w/\kappa_{\mathrm{cov}}\). More generally,
for linear and RKHS classes,
Appendix~\ref{app:weight-stability} reduces this condition to weighted
Gram-matrix and covariance-operator comparability, respectively.

We state the ERM rates in terms of metric-entropy critical radii. For
\(\varepsilon>0\), let
\(N(\varepsilon,\mathcal F,L^2(\mathsf Q))\) denote the
\(\varepsilon\)-covering number of \(\mathcal F\) under \(L^2(\mathsf Q)\), and
define
\[
\mathcal J(\delta,\mathcal F)
:=
\int_0^\delta
\sup_{\mathsf Q}
\sqrt{\log N(\varepsilon,\mathcal F,L^2(\mathsf Q))}\,d\varepsilon,
\quad
\delta_n
:=
\inf\left\{
\delta>0:
\mathcal J(\delta,\mathcal F)
\le
\sqrt n\,\delta^2
\right\}
\vee\sqrt{\frac{\log(en)}{n}},
\]
where the supremum is over finitely supported distributions \(\mathsf Q\).
The entropy component of \(\delta_n\) yields localized ERM rates for common
function classes: order \(\sqrt{s\log(p/s)/n}\) for sparse linear classes,
\(\sqrt{V\log(n/V)/n}\) for VC-subgraph classes, and
\(n^{-s/(2s+d)}\) for \(d\)-variate Hölder or Sobolev classes of
smoothness \(s\)
\citep{van1996weak,nickl2007bracketing,mendelson2010regularization,
wainwright2019high}.

We summarize the effects of weight estimation and sampling through
\[
\begin{aligned}
\varepsilon_{\mathrm{Bell},w}(Q)
:=
\left\|
\left(\frac{\widehat w_\beta}{w_\beta}-1\right)
(\mathcal TQ-\mathcal T_{\mathcal F,\beta}Q)
\right\|_{2,d_\beta}, \quad \varepsilon_{\mathrm{stat}}(m,\tau)
:=
\sqrt{\kappa_{\mathrm{cov}}}
\left(
\delta_n+\sqrt{\frac{\log(em/\tau)}{n}}
\right).
\end{aligned}
\]

\begin{lemma}[One-step weighted regression error]
\label{lemma::errorperiter}
Assume Conditions~\ref{cond::stationary}--\ref{cond::weight-stability}. There
exists \(C=C(M)<\infty\) such that, for any \(\tau\in(0,1)\), with probability
at least \(1-\tau\),
\[
\left\|
\widehat Q^{(k+1)}
-\mathcal T_{\mathcal F,\beta}\widehat Q^{(k)}
\right\|_{2,d_\beta}
\le
C\left\{
\frac{\varepsilon_{\mathrm{stat}}(1,\tau)}
{\sqrt{\lambda_{\mathcal F}}}
+
\frac{\varepsilon_{\mathrm{Bell},w}(\widehat Q^{(k)})}
{\lambda_{\mathcal F}}
\right\}.
\]
\end{lemma}

The first term is the finite-sample regression error. Weight-estimation error
enters only through the Bellman projection residual at the current iterate.
Writing
\(\varepsilon_w:=\|\widehat w_\beta/w_\beta-1\|_{2,d_\beta}\),
H\"older's inequality gives, for every \(Q\in\mathcal F\),
\[
\varepsilon_{\mathrm{Bell},w}(Q)
\le
\varepsilon_w
\|\mathcal TQ-\mathcal T_{\mathcal F,\beta}Q\|_\infty.
\]
Thus, weight error and Bellman projection error interact multiplicatively.
Exact occupancy weighting or Bellman completeness eliminates this term, while
approximate completeness reduces its sensitivity to weight-estimation error.

For unweighted FQE, \(\widehat w_\beta\equiv1\), so the relative weight error
is the fixed distribution mismatch \(w_\beta^{-1}-1\). The corresponding
term therefore persists unless the distribution mismatch or Bellman projection
residual is small, and vanishes under matched occupancy sampling or Bellman
completeness. Occupancy weighting instead replaces this fixed mismatch by
density-ratio estimation error, which can vanish with sample size.

\subsection{Main Finite-Sample Guarantee}
\label{sec:theory}

We next quantify two mechanisms for controlling
\(\varepsilon_{\mathrm{Bell},w}(\widehat Q^{(k)})\): completeness and
realizability. Define the value-function approximation error and inherent
Bellman error by
\[
\varepsilon_{\mathrm{app}}
:=
\inf_{f\in\mathcal F}\|f-Q^\pi\|_{2,d_\beta},
\qquad
\varepsilon_{\mathrm{Bell}}
:=
\sup_{Q\in\mathcal F}
\|\mathcal TQ-\mathcal T_{\mathcal F,\beta}Q\|_{2,d_\beta}.
\]
Small \(\varepsilon_{\mathrm{Bell}}\) controls the Bellman projection residual
uniformly over \(\mathcal F\). Realizability instead provides local control:
if \(Q^\pi\in\mathcal F\), then
\(\mathcal TQ^\pi-\mathcal T_{\mathcal F,\beta}Q^\pi=0\), so the
weight-error interaction vanishes at the true fixed point. To extend this
local control to approximate realizability, we impose the following
interpolation condition.

\begin{enumerate}[label=\textbf{C\arabic*}, ref={C\arabic*},
leftmargin=1.5em, resume=cond]
\item \label{cond::bellman-target-interp}
\textbf{Interpolation bound for Bellman projection residuals.}
There exist \(\alpha\in[0,1]\) and \(C_\infty<\infty\) such that, for every
\(Q\in\mathcal F\),
\[
\|\mathcal TQ-\mathcal T_{\mathcal F,\beta}Q\|_\infty
\le
C_\infty
\|\mathcal TQ-\mathcal T_{\mathcal F,\beta}Q\|_{2,d_\beta}^{\alpha}.
\]
\end{enumerate}
Condition~\ref{cond::bellman-target-interp} rules out projection residuals
that are small in \(L^2(d_\beta)\) but large on narrow regions. At
\(\alpha=0\), with the convention \(0^0=1\), one may always take
\[
C_\infty
:=
\sup_{Q\in\mathcal F}
\|\mathcal TQ-\mathcal T_{\mathcal F,\beta}Q\|_\infty,
\]
which is finite under Condition~\ref{cond::bounded}. Under the sufficient conditions in Appendix~\ref{appendix:supnorm}, the
interpolation inequality holds with exponent \(1\) for finite-dimensional
linear classes, with \(\alpha=1-1/(2r)\) for RKHS classes whose eigenvalues
satisfy \(\lambda_j\lesssim j^{-2r}\) \citep{mendelson2010regularization}, \(r>1/2\), with
\(\alpha=2s/(2s+d)\) for H\"older classes of smoothness \(s>0\) \citep{triebel2006theory}, and with
\(\alpha=1-d/(2s)\) for Sobolev classes of smoothness \(s>d/2\) on
\(d\)-dimensional domains \citep{adams2003sobolev}. The following theorem quantifies the distinct roles
of completeness and realizability in controlling the weight-error interaction.

\begin{theorem}[Finite-sample error bound]
\label{thm:completeness-realizability}
Assume Conditions~\ref{cond::stationary}--\ref{cond::weight-stability} and
Condition~\ref{cond::bellman-target-interp} with \(\alpha\in[0,1)\). Let
\(\gamma_{\beta,\star}:=(1+\bar\gamma_\beta)/2\). There exists
\(C=C(M,\alpha,\gamma,\beta)<\infty\) such that, for every
\(\tau\in(0,1)\) and \(K\in\mathbb N\), with probability at least
\(1-\tau\),
\[
\begin{aligned}
\|\widehat Q^{(K)}-Q^\pi\|_{2,d_\beta}
&\le
\gamma_{\beta,\star}^K
\|\widehat Q^{(0)}-Q^\pi_{\mathcal F,\beta}\|_{2,d_\beta}
+\frac{\varepsilon_{\mathrm{app}}}{1-\bar\gamma_\beta}
+\frac{C}{1-\gamma_{\beta,\star}}
\frac{\varepsilon_{\mathrm{stat}}(K,\tau)}
{\sqrt{\lambda_{\mathcal F}}}
\\
&\quad+
\frac{C}{1-\gamma_{\beta,\star}}
\min\left\{
\frac{C_\infty\varepsilon_w}{\lambda_{\mathcal F}}
\varepsilon_{\mathrm{Bell}}^\alpha,
\left(\frac{C_\infty\varepsilon_w}{\lambda_{\mathcal F}}\right)^{1/(1-\alpha)}
+\frac{C_\infty\varepsilon_w}{\lambda_{\mathcal F}}
\left(\frac{\varepsilon_{\mathrm{app}}}{1-\bar\gamma_\beta}\right)^\alpha
\right\}.
\end{aligned}
\]
\end{theorem}

At \(\alpha=0\), the bound gives the generic linear dependence on
\(C_\infty\varepsilon_w/\lambda_{\mathcal F}\), available under
Condition~\ref{cond::bounded}. For \(\alpha\in(0,1)\), completeness and
realizability control the weight-error term in complementary ways: exact
Bellman completeness eliminates it, while exact realizability leaves the
higher-order term
\((C_\infty\varepsilon_w/\lambda_{\mathcal F})^{1/(1-\alpha)}\).
For linear function approximation,
Corollary~\ref{cor:linear-endpoint-refinement} in
Appendix~\ref{appendix:supnorm} treats the endpoint \(\alpha=1\), where the
weight-error term becomes
\((\varepsilon_w/\lambda_{\mathcal F})
\min\{\varepsilon_{\mathrm{Bell}},
\varepsilon_{\mathrm{app}}/(1-\bar\gamma_\beta)\}\)
and the effective contraction factor is
\(\gamma_w=\bar\gamma_\beta
(1+C\varepsilon_w/\lambda_{\mathcal F})<1\).

\textbf{End-to-end guarantee for FQE with FORE weights.}
We next analyze the FORE estimator in
Algorithm~\ref{alg:fore-weighted-fqe} \citep{van2026fitted}. FORE controls
estimation error in generalized KL divergence,
\[
D_{\nu_b}^{\rm gen}(f\|g)
:=
\E_{\nu_b}[f\log(f/g)-f+g],
\]
whereas the FQE bound depends on relative \(L^2(d_\beta)\) error. The
following condition allows us to translate between these two error measures.

\begin{enumerate}[label=\textbf{C\arabic*}, ref={C\arabic*},
leftmargin=1.5em, resume=cond]
\item \label{cond::fore-target-ratio-bounded}
\textbf{Target-ratio boundedness.}
The target occupancy ratio satisfies
\[
0<\underline\kappa_\beta
:=
\operatorname*{ess\,inf}_{x\sim\nu_b}w_\beta(x)
\le
\operatorname*{ess\,sup}_{x\sim\nu_b}w_\beta(x)
=:
\overline\kappa_\beta
<\infty.
\]
\end{enumerate}

The finite-sample FORE rate depends on the occupancy contraction coefficient
\(\beta\), the critical radius \(\mathfrak r_{n_w,\rm fit}\), and the
centered log-ratio bound \(R_w\), which implies
\(e^{-2R_w}\le\widehat w^{(k)}\le e^{2R_w}\).
Appendix~\ref{app:fore-estimator}--\ref{app:fore-conditions} gives the precise
definitions and conditions. Let \(\varrho_\beta:=(1+\beta)/2\). Then
Theorem~\ref{thm:fore-stationary-weight} implies that, with probability at
least \(1-\tau\),
\(\varepsilon_w\le\varepsilon_{\rm FORE}(K_w,n_w,\tau)\), where
\begin{equation}
\label{eq:main-fore-radius}
\begin{aligned}
\varepsilon_{\rm FORE}(K_w,n_w,\tau)
:=C_{{\rm FORE},\beta}^{\circ}\Bigg[
&\frac{\varrho_\beta^{K_w/2}}
{\sqrt{\underline\kappa_\beta}}
\{D_{\nu_b}^{\rm gen}(1\|w_\beta)\}^{1/2}
+
\frac{1}{\underline\kappa_\beta}
\left\{
\frac{\inf_{w\in\mathcal W_{\rm F}}
D_{\nu_b}^{\rm gen}(w\|w_\beta)}
{1-\beta}
\right\}^{1/2}
\\[-0.25em]
&\quad+
\frac{\log(e n_w)}
{(1-\beta)\sqrt{\underline\kappa_\beta}}
\left\{
\mathfrak r_{n_w,\rm fit}
+
\sqrt{\frac{\log(1/\tau)}{n_w}}
\right\}
\Bigg],
\end{aligned}
\end{equation}
for a finite constant \(C_{{\rm FORE},\beta}^{\circ}\) depending only on the
fixed constants in the FORE conditions. The three terms correspond to
finite-iteration, ratio-class approximation, and statistical error,
respectively. Substituting this rate into
Theorem~\ref{thm:completeness-realizability} yields the following end-to-end
guarantee.

\begin{corollary}[End-to-end guarantee for FQE with FORE weights]
\label{cor:fore-fqe-end-to-end}
Assume Conditions~\ref{cond::stationary}--\ref{cond::split},
Condition~\ref{cond::bellman-target-interp} with \(\alpha\in[0,1)\),
Condition~\ref{cond::fore-target-ratio-bounded}, and
Conditions~\ref{cond::fore-sample}--\ref{cond::fore-class}.
Let \((\widehat w_\beta,\widehat Q^{(K)})\) be the output of
Algorithm~\ref{alg:fore-weighted-fqe}. Then
Conditions~\ref{cond::weight-coverage} and~\ref{cond::weight-stability} hold
with
\[
\kappa_{\rm cov}
=
\overline\kappa_{\rm F}
:=
\max\{\overline\kappa_\beta,e^{2R_w}\},
\qquad
\lambda_{\mathcal F}
\ge
\frac{e^{-2R_w}}{\overline\kappa_\beta}.
\]
Moreover, for every \(\tau\in(0,1)\), with probability at least \(1-\tau\),
the conclusion of Theorem~\ref{thm:completeness-realizability} holds with
\(\varepsilon_w\) replaced by
\(\varepsilon_{\rm FORE}(K_w,n_w,\tau/2)\) and
\(\varepsilon_{\rm stat}(K,\tau)\) replaced by
\(\varepsilon_{\rm stat}(K,\tau/2)\), for a constant
\(C=C(M,C_\infty,\alpha,\beta,\gamma,R_w,
\underline\kappa_\beta^{-1},\overline\kappa_\beta)\).
\end{corollary}

Condition~\ref{cond::fore-target-ratio-bounded} has a simple sufficient
condition for its lower bound. Let \(\pi_b(\cdot\mid s)\) denote the
conditional action law under \(\nu_b\), and define
\(\underline\omega_{\pi/b}
:=\operatorname*{ess\,inf}_{(s,a)\sim\nu_b}
\pi(a\mid s)/\pi_b(a\mid s)\).
If \(\underline\omega_{\pi/b}>0\), the occupancy recursion implies
\(\underline\kappa_\beta
\ge(1-\beta)\underline\omega_{\pi/b}\), and hence
\(\underline\kappa_\beta^{-1}
\le\{(1-\beta)\underline\omega_{\pi/b}\}^{-1}\).
Thus, under uniform action overlap, the worst-case inverse lower bound grows
at most linearly with the effective horizon \((1-\beta)^{-1}\). This rate
uses only the restart component of the occupancy recursion; the actual
occupancy ratio may be bounded further away from zero.

Under joint realizability of \(w_\beta\) and \(Q^\pi\), the ratio- and
value-class approximation errors vanish. The remaining bound contains only
finite-iteration and statistical terms, even when the Bellman projection
residual is nonzero away from \(Q^\pi\). In particular, the FORE contribution
is of order
\(\{\varepsilon_{\rm FORE}(K_w,n_w,\tau/2)/
\lambda_{\mathcal F}\}^{1/(1-\alpha)}\), which is higher order in the
weight-error radius whenever \(\alpha>0\). Thus, as the iteration errors and
statistical radii vanish, joint realizability suffices for consistency without
Bellman completeness.

\textbf{State-value error.}
Let \(d_\beta^S\) denote the state marginal of \(d_\beta\). Since
\(d_\beta(da\mid s)=\pi(da\mid s)\), Jensen's inequality gives
\(\|\pi\widehat Q-V^\pi\|_{2,d_\beta^S}
\le \|\widehat Q-Q^\pi\|_{2,d_\beta}\).
For the offline state marginal, define
\(\kappa_{\beta,S}:=\inf\{\kappa\ge1:\nu_b^S\le\kappa d_\beta^S\}\), so
\(\|\pi\widehat Q-V^\pi\|_{2,\nu_b^S}\le
\sqrt{\kappa_{\beta,S}}\|\widehat Q-Q^\pi\|_{2,d_\beta}\).
The restart construction implies
\(\kappa_{\beta,S}\le(1-\beta)^{-1}\), so
\((1-\beta)^{-1/2}\) is only a worst-case transfer factor and the bound
adapts to the actual state-marginal overlap. By Cauchy--Schwarz, the same
bounds control the corresponding average value errors.
Appendix~\ref{app:discounted-occupancy-fqe} treats other initial-state
distributions.

\section{Experimental Investigation}
\label{sec:experiments}

We study two settings that isolate the mechanisms identified by the theory.
First, finite-state examples examine instability from mismatch between the
offline regression distribution and the target-policy occupancy distribution.
Second, a stochastic continuing-control experiment evaluates FQE with learned
occupancy weights. Appendix~\ref{app:toy_norm_mismatch} and
Appendix~\ref{app:cartpole-study} give additional experimental details
and sensitivity analyses.

\begin{figure}[t]
    \centering
    \includegraphics[width=\linewidth]
    {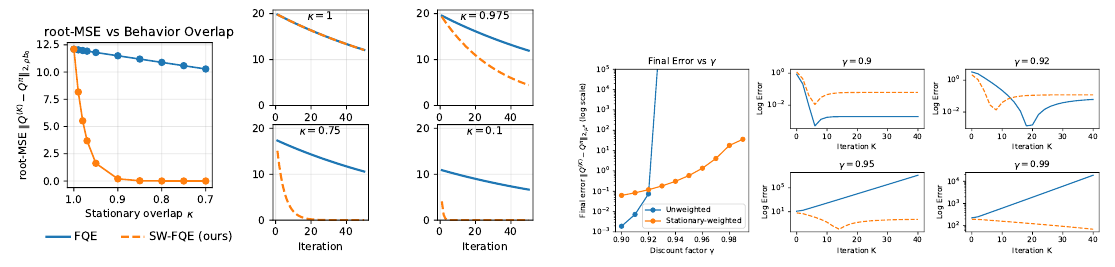}
    \caption{\textbf{Finite-state norm mismatch.}
    Left: occupancy weighting restores rapid error decay in Baird's example
    as behavior--target overlap decreases.
    Right: occupancy weighting prevents the high-discount divergence observed
    in sparse Garnet MDPs.}
    \label{fig:finite-state-norm-mismatch}
\end{figure}

\noindent\textbf{Finite-state norm mismatch.}\space
We consider two finite-state settings in which the projection geometry can be
computed directly: an adaptation of Baird's seven-state example
\citep{baird1995residual} and sparse Garnet MDPs
\citep{bhatnagar2009natural}. With exact occupancy weighting, FQE remains
stable as behavior--target mismatch increases, whereas unweighted FQE
converges slowly or diverges. Figure~\ref{fig:finite-state-norm-mismatch}
shows that reweighting the Bellman regression toward the target-policy
occupancy restores the contraction predicted by the theory.

\noindent\textbf{Stochastic continuing CartPole with learned weights.}\space
We next consider a stochastic continuing-reset CartPole-inspired MDP. Failure
induces a new state through the transition kernel, so Bellman targets continue
across resets. The behavior and target policies are full-support logistic
policies. Each offline dataset contains \(25{,}000\) transitions sampled from
the stationary behavior distribution after burn-in. Unweighted FQE and
FORE-weighted FQE use the same rewards, transitions, function class, and
initialization; FORE is estimated without rewards and changes only the
regression weights. We set the occupancy and Bellman discounts to
\(\beta=\gamma=0.98\) and use the offline state marginal as the restart
distribution.

We evaluate the two estimators on \(\CartPoleN\) independent
offline datasets. We measure absolute target-policy value error using an
independently simulated Monte Carlo value as the reference. We report the
paired mean reduction in absolute error and its 95\% bootstrap confidence
interval, together with the effective sample size of the estimated FORE
weights.

\begin{figure}[t]
  \centering
  \includegraphics[width=0.62\linewidth]
  {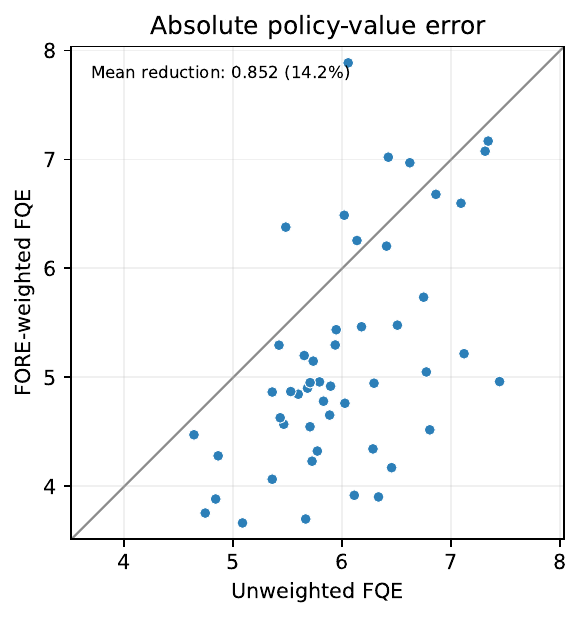}
  \caption{\textbf{FORE-weighted FQE in stochastic continuing CartPole.}
  Each point gives the absolute target-policy value error for one offline
  dataset. The diagonal marks equal error; points below it favor FORE-weighted
  FQE.}
  \label{fig:cartpole-fore-fqe}
\end{figure}

\noindent\textbf{Learned-weight results.}\space
FORE-FQE reduces mean absolute value error from
\(\CartPoleUnweightedMAE\) to \(\CartPoleFOREMAE\), corresponding to a mean
relative reduction of \(\CartPoleRelativeReduction\). The paired mean error
reduction has 95\% bootstrap confidence interval
\([\CartPoleGainCILower,\CartPoleGainCIUpper]\), and FORE-FQE improves on
unweighted FQE in \(\CartPoleFOREWins\) of the
\(\CartPoleN\) datasets. The median effective sample size of the
FORE weights, relative to the offline sample size, is
\(\CartPoleMedianESS\).

\section{Discussion}

We studied the instability of off-policy FQE through the mismatch between the
offline regression norm and the target-policy contraction geometry.
Occupancy-weighted FQE changes only the regression weights so that projection
occurs in a target-policy occupancy norm, restoring contraction while
preserving the fitted-regression structure of FQE.

Our finite-sample analysis shows that density-ratio estimation error enters
through its interaction with the Bellman projection residual. Approximate
Bellman completeness and value-function realizability control this interaction
uniformly and locally, respectively, while exact realizability yields
higher-order dependence on ratio-estimation error. Combined with fitted
occupancy-ratio evaluation, joint realizability of the value function and
occupancy ratio suffices for consistent estimation under coverage, without
Bellman or critic-side completeness.

Occupancy weighting does not eliminate the need for adequate coverage or
reliable estimation of the occupancy weights. Our bounds nevertheless show
that global density-ratio accuracy can be stronger than necessary: weight
estimation error matters through its interaction with the fitted Bellman
residual. This also suggests a role for regularizing estimated occupancy
weights in finite samples. For example, one may temper and renormalize the
estimated ratio,
\[
\widetilde w_\alpha(x)
\propto
\widehat w_\beta(x)^\alpha,
\qquad
\alpha\in[0,1],
\]
to reduce variability from extreme weights
\citep{sinha2022experience}. Tempering trades exact occupancy alignment for
more stable regression and remains compatible with the contraction argument
whenever the resulting weighted Bellman update is contractive.

\bibliographystyle{plainnat}
\bibliography{ref}

\appendix

\appendix

\setcounter{equation}{0}
\renewcommand{\theequation}{S\arabic{equation}}
\setcounter{theorem}{0}
\setcounter{figure}{0}
\setcounter{table}{0}
\setcounter{lemma}{0}
\setcounter{corollary}{0}
\renewcommand{\thetheorem}{S\arabic{theorem}}
\renewcommand{\thecorollary}{S\arabic{corollary}}
\renewcommand{\thelemma}{S\arabic{lemma}}
\renewcommand{\thefigure}{S\arabic{figure}}
\renewcommand{\thetable}{S\arabic{table}}
\renewcommand{\thealgorithm}{S\arabic{algorithm}}

\paragraph{Appendix roadmap.}
Appendix~\ref{app:discounted-occupancy-fqe} develops
FQE with discounted-occupancy weights and doubly robust discounted-value
estimation.
Appendix~\ref{app:additional-related-work} provides additional related work.
Appendix~\ref{app:additional-details} collects the occupancy-weighted algorithm,
interpolation bounds for weight--residual interactions, approximate-weight
analysis, and the implicit stationary-sampling extension.
Appendix~\ref{sec::appendix} gives details for the finite-state and CartPole
experiments, and
Appendix~\ref{appendix::stationary} gives the contraction results and proofs
for the discounted-occupancy analysis.
Appendices~\ref{app:technical-lemmas} and~\ref{app:main-proofs} contain the
technical lemmas and proofs of the main results. Finally,
Appendix~\ref{app:fore-weight-bound} gives the weight-error bound for FORE and the
resulting end-to-end FQE guarantee.

\section{FQE with Discounted-Occupancy Weights and Doubly Robust Value Estimation}
\label{app:discounted-occupancy-fqe}

The main construction takes \(\rho=\nu_b^S\). This appendix allows a general
initial-state distribution \(\rho\) and develops the corresponding
population and empirical FQE recursions and discounted-value estimator. We also
establish two robustness
properties: the doubly robust estimator is valid if either the occupancy ratio
or the value function is correctly specified, and, with an unrestricted
intercept, the plug-in value at the oracle projected fixed point remains valid
under \(Q\)-function misspecification. Proofs are given in
Appendix~\ref{app:discounted-occupancy-proofs}.

\subsection{Discounted occupancy and population recursion}

Fix an initial state distribution \(\rho\), let
\(\xi_0:=\rho\otimes\pi\) denote the induced initial state-action distribution,
and let \(\gamma'\in(0,1)\) be an occupancy discount, possibly different from
the Bellman discount \(\gamma\). Define
\begin{equation}
\label{eq:discounted-occupancy}
d_{\pi,\gamma'}
:=
(1-\gamma')\sum_{t=0}^{\infty}(\gamma')^t\xi_0P_\pi^t.
\end{equation}
This distribution satisfies
\begin{equation}
\label{eq:discounted-balance}
d_{\pi,\gamma'}
=
(1-\gamma')\xi_0+\gamma'd_{\pi,\gamma'}P_\pi.
\end{equation}
When \(d_{\pi,\gamma'}\ll\nu_b\), define the corresponding density ratio
\[
w_{\pi,\gamma'}
:=
\frac{\mathrm d d_{\pi,\gamma'}}{\mathrm d\nu_b}.
\]

For a function class \(\mathcal F\), define
\[
\Pi_{\mathcal F,\gamma'}g
:=
\argmin_{f\in\mathcal F}\|f-g\|_{2,d_{\pi,\gamma'}},
\qquad
\mathcal T_{\mathcal F,\gamma'}
:=
\Pi_{\mathcal F,\gamma'}\mathcal T.
\]
We impose the following discounted analogues of
Conditions~\ref{cond::stationary}--\ref{cond::convex}.

\begin{enumerate}[label=\textbf{D\arabic*}, ref={D\arabic*},
leftmargin=1.5em, series=disccond]
\item \label{cond:discounted-coverage}
\textbf{Discounted occupancy and coverage.}
The distribution \(d_{\pi,\gamma'}\) satisfies
\(d_{\pi,\gamma'}\ll\nu_b\), and
\(r_0\in L^2(d_{\pi,\gamma'})\).

\item \label{cond:discounted-convex}
\textbf{Closed convex class.}
The class \(\mathcal F\) is nonempty, closed, and convex in
\(L^2(d_{\pi,\gamma'})\).
\end{enumerate}

Condition~\ref{cond:discounted-coverage} requires coverage only of the
discounted occupancy induced by \((\rho,\pi)\), not existence of a stationary
distribution.

\begin{theorem}[Contraction after projection in the discounted-occupancy norm]
\label{thm:discounted-projected-contraction}
Suppose Conditions~\ref{cond:discounted-coverage}--%
\ref{cond:discounted-convex} hold and \(\gamma'>\gamma^2\). Define
\[
\bar\gamma_{\gamma'}
:=
\frac{\gamma}{\sqrt{\gamma'}}
<1.
\]
Then, for all \(Q,Q'\in\mathcal F\),
\[
\|\mathcal T_{\mathcal F,\gamma'}Q
-\mathcal T_{\mathcal F,\gamma'}Q'\|_{2,d_{\pi,\gamma'}}
\le
\bar\gamma_{\gamma'}
\|Q-Q'\|_{2,d_{\pi,\gamma'}}.
\]
Consequently, \(\mathcal T_{\mathcal F,\gamma'}\) has a unique fixed point
\(Q^\pi_{\mathcal F,\gamma'}\), and
\(Q_{\mathcal F,\gamma'}^{(k+1)}
=\mathcal T_{\mathcal F,\gamma'}Q_{\mathcal F,\gamma'}^{(k)}\) satisfies
\begin{align}
\label{eq:discounted-population-recursion}
\|Q_{\mathcal F,\gamma'}^{(K)}
-Q^\pi_{\mathcal F,\gamma'}\|_{2,d_{\pi,\gamma'}}
&\le
\bar\gamma_{\gamma'}^K
\|Q_{\mathcal F,\gamma'}^{(0)}
-Q^\pi_{\mathcal F,\gamma'}\|_{2,d_{\pi,\gamma'}},
\\
\label{eq:discounted-approximation-bound}
\|Q^\pi_{\mathcal F,\gamma'}-Q^\pi\|_{2,d_{\pi,\gamma'}}
&\le
\frac{1}{1-\bar\gamma_{\gamma'}}
\inf_{f\in\mathcal F}
\|f-Q^\pi\|_{2,d_{\pi,\gamma'}}.
\end{align}
In particular, \(\gamma'=\gamma\) yields contraction modulus
\(\sqrt{\gamma}\). Taking \(\rho=\nu_b^S\) and \(\gamma'=\beta\) recovers
the main-text contraction.
\end{theorem}

\begin{lemma}[Inexact recursion under discounted-occupancy weighting]
\label{lem:discounted-inexact-recursion}
Under the conditions of Theorem~\ref{thm:discounted-projected-contraction},
suppose \(\widehat Q^{(k)}\in\mathcal F\) and
\[
\|\widehat Q^{(k)}
-\mathcal T_{\mathcal F,\gamma'}\widehat Q^{(k-1)}
\|_{2,d_{\pi,\gamma'}}
\le
\eta_{k,\gamma'}.
\]
Then
\begin{equation}
\label{eq:discounted-inexact-recursion}
\|\widehat Q^{(K)}-Q^\pi_{\mathcal F,\gamma'}
\|_{2,d_{\pi,\gamma'}}
\le
\bar\gamma_{\gamma'}^K
\|\widehat Q^{(0)}-Q^\pi_{\mathcal F,\gamma'}
\|_{2,d_{\pi,\gamma'}}
+
\sum_{j=1}^K
\bar\gamma_{\gamma'}^{K-j}\eta_{j,\gamma'}.
\end{equation}
Hence, the one-step regression analysis in the main text applies with
\((d_\beta,w_\beta,\bar\gamma_\beta)\) replaced by
\((d_{\pi,\gamma'},w_{\pi,\gamma'},\bar\gamma_{\gamma'})\). A uniform
one-step error is amplified by at most
\((1-\bar\gamma_{\gamma'})^{-1}\).
\end{lemma}

\subsection{Algorithm}

The discounted-occupancy projection is implemented by reweighting the Bellman
regressions of ordinary FQE. The occupancy ratio is estimated once and reused
across iterations.

\begin{algorithm}[H]
\caption{\textbf{Fitted \(Q\)-Evaluation with Discounted-Occupancy Weights}}
\label{alg:discounted-occupancy-fqe}
\begin{algorithmic}[1]
\INPUT Offline data \(\mathcal D_n\), target policy \(\pi\), initial
distribution \(\rho\), function class \(\mathcal F\), iterations \(K\),
Bellman discount \(\gamma\), occupancy discount \(\gamma'>\gamma^2\)
\STATE Estimate a nonnegative ratio
\(\widehat w_{\pi,\gamma'}\approx
\mathrm d d_{\pi,\gamma'}/\mathrm d\nu_b\)
\STATE Initialize \(\widehat Q^{(0)}\in\mathcal F\)
\FOR{\(k=0,\ldots,K-1\)}
\STATE
\(
\displaystyle
\widehat Q^{(k+1)}
\in
\argmin_{f\in\mathcal F}
\frac1n\sum_{i=1}^n
\widehat w_{\pi,\gamma'}(S_i,A_i)
\{R_i+\gamma(\pi\widehat Q^{(k)})(S_i')-f(S_i,A_i)\}^2
\)
\ENDFOR
\OUTPUT \(\widehat Q^{(K)}\)
\end{algorithmic}
\end{algorithm}

With oracle weights, the population target of each regression is
\(\mathcal T_{\mathcal F,\gamma'}\widehat Q^{(k)}\). Weight-estimation and
sampling errors therefore enter through the one-step errors
\(\eta_{k,\gamma'}\) in Lemma~\ref{lem:discounted-inexact-recursion}. For
discounted policy-value estimation, we set \(\gamma'=\gamma\), aligning the
regression norm with the occupancy distribution that defines the value.

\subsection{Doubly robust discounted-value estimation}

Set \(\gamma'=\gamma\), abbreviate
\(d_\gamma:=d_{\pi,\gamma}\) and \(w_\gamma:=w_{\pi,\gamma}\), and define the
normalized discounted value
\begin{equation}
\label{eq:normalized-discounted-value}
J_\gamma(\pi)
:=
(1-\gamma)\E_{S_0\sim\rho}[(\pi Q^\pi)(S_0)]
=
\E_{d_\gamma}[r_0(S,A)].
\end{equation}
For any candidate ratio \(w\) and value function \(Q\), define
\begin{equation}
\label{eq:dr-population-functional}
\Psi_{\rm DR}(w,Q)
:=
(1-\gamma)\E_{S_0\sim\rho}[(\pi Q)(S_0)]
+
\E_{\nu_b}\!\left[
w(S,A)\{\mathcal TQ(S,A)-Q(S,A)\}
\right].
\end{equation}

\begin{theorem}[Double robustness and product remainder]
\label{thm:discounted-dr-identity}
Suppose \(d_\gamma\ll\nu_b\) and the displayed expectations are finite. Then
\begin{equation}
\label{eq:dr-exact-remainder}
\Psi_{\rm DR}(w,Q)-J_\gamma(\pi)
=
\E_{\nu_b}\!\left[
\{w-w_\gamma\}(\mathcal TQ-Q)
\right].
\end{equation}
Hence, \(\Psi_{\rm DR}(w,Q)=J_\gamma(\pi)\) if either
\(w=w_\gamma\) \(\nu_b\)-almost surely or \(Q=Q^\pi\).

Define
\[
\varepsilon_w(w)
:=
\left\|
\frac{w-w_\gamma}{\sqrt{w_\gamma}}
\right\|_{2,\nu_b},
\]
with the convention that this quantity is infinite unless
\(w=w_\gamma\) on \(\{w_\gamma=0\}\). Then
\begin{align}
\label{eq:dr-product-bound}
|\Psi_{\rm DR}(w,Q)-J_\gamma(\pi)|
&\le
\varepsilon_w(w)\|\mathcal TQ-Q\|_{2,d_\gamma}
\\
&\le
(1+\sqrt\gamma)\varepsilon_w(w)
\|Q-Q^\pi\|_{2,d_\gamma}.
\nonumber
\end{align}
Thus, the bias is controlled by the product of ratio and value-function
errors.
\end{theorem}

For a sample-split estimator, let \(\widehat w\) and \(\widehat Q\) be fitted
on an auxiliary sample. On an independent evaluation sample, observe
\(O_i=(S_i,A_i,R_i,S_i')\), with \((S_i,A_i)\sim\nu_b\), and independent
\(S_{0i}\sim\rho\). Define
\begin{equation}
\label{eq:dr-estimator}
\widehat J_{\rm DR}
:=
\frac1n\sum_{i=1}^n
\left[
(1-\gamma)(\pi\widehat Q)(S_{0i})
+\widehat w(S_i,A_i)
\{R_i+\gamma(\pi\widehat Q)(S_i')-\widehat Q(S_i,A_i)\}
\right].
\end{equation}
If the expectation under \(\rho\) is known, it may replace the first empirical
average. Cross-fitting applies \eqref{eq:dr-estimator} foldwise and averages
the resulting estimates
\citep{jiang2016doubly,kallus2020double,kallus2022efficiently,
van2025automaticDRL}.

\begin{corollary}[Finite-sample bound for the cross-fitted doubly robust estimator]
\label{cor:discounted-drl-bound}
Condition on the auxiliary sample and suppose the summands
\(\widehat\psi_i\) in \eqref{eq:dr-estimator} are i.i.d. with conditional
variance at most \(\sigma_{\rm DR}^2\). Then, for every \(\tau\in(0,1)\),
with conditional probability at least \(1-\tau\),
\begin{equation}
\label{eq:drl-finite-sample-bound}
|\widehat J_{\rm DR}-J_\gamma(\pi)|
\le
\frac{\sigma_{\rm DR}}{\sqrt{n\tau}}
+
(1+\sqrt\gamma)\varepsilon_w(\widehat w)
\|\widehat Q-Q^\pi\|_{2,d_\gamma}.
\end{equation}
If the centered score is conditionally \(\sigma_{\rm DR}\)-sub-Gaussian, the
first term may instead be replaced by
\(\sigma_{\rm DR}\sqrt{2\log(2/\tau)/n}\).

If \(\widehat Q=\widehat Q^{(K)}\) is produced by
Algorithm~\ref{alg:discounted-occupancy-fqe} with \(\gamma'=\gamma\), then
\begin{align}
\label{eq:drl-fqe-product-bound}
|\widehat J_{\rm DR}-J_\gamma(\pi)|
&\le
\frac{\sigma_{\rm DR}}{\sqrt{n\tau}}
+(1+\sqrt\gamma)\varepsilon_w(\widehat w)
\Bigg[
\gamma^{K/2}
\|\widehat Q^{(0)}-Q^\pi_{\mathcal F,\gamma}
\|_{2,d_\gamma}
\nonumber\\
&\qquad
+\sum_{j=1}^K\gamma^{(K-j)/2}\eta_{j,\gamma}
+\frac{1}{1-\sqrt\gamma}
\inf_{f\in\mathcal F}\|f-Q^\pi\|_{2,d_\gamma}
\Bigg].
\end{align}
\end{corollary}

The first term in \eqref{eq:drl-fqe-product-bound} is evaluation noise. All
FQE errors are multiplied by the ratio error. Any bound
\(\varepsilon_w(\widehat w)\le\mathcal E_w\), including one obtained from
discounted FORE \citep{van2026fitted}, can therefore be substituted directly
into \eqref{eq:drl-fqe-product-bound}.

Under cross-fitting, suppose the fitted score converges in \(L^2\) to the
discounted-OPE influence function, its empirical average satisfies a central
limit theorem, and
\(\varepsilon_w(\widehat w)
\|\widehat Q-Q^\pi\|_{2,d_\gamma}=o_p(n^{-1/2})\),
so that the product remainder is first-order negligible. The estimator is then
asymptotically linear and attains the corresponding semiparametric efficiency
bound
\citep{kallus2020double,kallus2022efficiently,van2025automaticDRL}.

\subsection{Value correctness under an unrestricted intercept}

The doubly robust identity yields a simple consequence when the fitted class
has an unrestricted intercept. At the oracle projected fixed point, the
intercept first-order condition forces the mean Bellman residual under the
discounted occupancy to vanish.

\begin{corollary}[Value correctness from an unrestricted intercept]
\label{cor:discounted-intercept}
Suppose Conditions~\ref{cond:discounted-coverage}--%
\ref{cond:discounted-convex} hold with \(\gamma'=\gamma\). Assume that, for
some \(c_0>0\),
\[
Q^\pi_{\mathcal F,\gamma}+c\in\mathcal F
\qquad
\text{for every }|c|<c_0.
\]
Then
\[
\E_{d_\gamma}
\left[
\mathcal TQ^\pi_{\mathcal F,\gamma}
-
Q^\pi_{\mathcal F,\gamma}
\right]
=0,
\]
and hence
\begin{equation}
\label{eq:intercept-value-correctness}
(1-\gamma)\E_{S_0\sim\rho}
[(\pi Q^\pi_{\mathcal F,\gamma})(S_0)]
=
J_\gamma(\pi).
\end{equation}
Thus, oracle occupancy weighting recovers the discounted policy value even
when \(Q^\pi_{\mathcal F,\gamma}\ne Q^\pi\).
\end{corollary}

Linear and affine classes with an unpenalized constant feature satisfy the
local intercept condition. The same mechanism has an empirical analogue: at
an exactly solved weighted empirical fixed point with an unrestricted
intercept, the empirical weighted Bellman residual has mean zero. If the
doubly robust estimator \eqref{eq:dr-estimator} is evaluated on the same
sample, its augmentation term therefore vanishes and the estimator reduces to
the plug-in value.

\section{Additional related work}
\label{app:additional-related-work}
\label{app:relatedwork}

\paragraph{Projected fixed points under linear function approximation.}

For fixed-policy evaluation with a linear value-function class, TD and LSTD
seek solutions to a projected Bellman equation, whereas exact population
least-squares FQE iterates the associated projected Bellman map
\citep{bradtke1996linear,tsitsiklis1996analysis,nedic2003least}.
Under an arbitrary off-policy projection norm, this map need not be
contractive, and its fixed point can be arbitrarily inaccurate even when
\(Q^\pi\) is well approximated by the class
\citep{yu2010error,kolter2011fixed}. Linearity alone therefore does not resolve
the geometric mismatch induced by off-policy projection. Existing convergence
guarantees require additional structure, such as Bellman completeness, linear-MDP
assumptions, or feature and sampling conditions that restore stability on the
chosen representation
\citep{melo2007convergence,wang2021statistical,chang2022learning,
jin2023provably}.

Related difficulties arise in control, where the policy depends on the current
\(Q\)-function and the projected Bellman optimality map is nonlinear. Under a
specialized adaptive training policy, \citet{meyn2024projected} established
boundedness of linear \(Q\)-learning iterates and existence of a projected
Bellman solution, while leaving convergence and uniqueness unresolved.
\citet{liu2025linear} subsequently obtained an \(L^2\) convergence rate to a
bounded set rather than to a particular projected solution. More recently,
\citet{mehta2026optimistic} constructed realizable examples with multiple
projected Bellman solutions, showing that additional structure is needed for
uniqueness and convergence. Together, these results demonstrate that linear
function approximation alone does not ensure stable projected Bellman
iteration. Our work instead concerns fixed-policy evaluation and uses occupancy
weighting to make the fitted-regression updates contractive for flexible,
possibly nonlinear function classes.

\section{Additional Details}
\label{app:additional-details}

\subsection{Algorithm}

\begin{algorithm}[h]
\caption{\textbf{Occupancy-Weighted Fitted \(Q\)-Evaluation}}
\label{alg::sw-fqi}
\begin{algorithmic}[1]

\INPUT Offline data \(\mathcal D_n\), target policy \(\pi\),
function class \(\mathcal F\), iterations \(K\), discounts \(\gamma,\beta\)

\ALGCOMMENT{Estimate occupancy ratio}
\STATE \(\widehat w_\beta \gets\) estimator of \(w_\beta\)

\STATE Initialize \(\widehat Q^{(0)} \in \mathcal F\)

\FOR{\(k = 0,\dots,K-1\)}

  \ALGCOMMENT{Construct Bellman targets}
  \STATE \(Y_i^{(k)} \gets R_i + \gamma(\pi \widehat Q^{(k)})(S_i')\)

  \ALGCOMMENT{Occupancy-weighted regression}
  \STATE \mbox{\(\widehat Q^{(k+1)} \gets
     \arg\min_{f \in \mathcal F}
     \sum_{i=1}^n \widehat w_\beta(S_i,A_i)
       \bigl\{Y_i^{(k)} - f(S_i,A_i)\bigr\}^2\)}

\ENDFOR

\OUTPUT \(\widehat Q^{(K)}\)

\end{algorithmic}
\end{algorithm}

\begin{figure}[t]
\centering
\begin{tikzpicture}[
    >=stealth,
    font=\small,
    every node/.style={inner sep=1pt}
]

\draw[thick, blue!60] (0,0) ellipse (3.0 and 1.6);
\node[blue!70!black] at (2.4,-1.4) {$L^2(d_\beta)$};

\draw[thick, red!70, rotate=8] (0,-0.05) ellipse (2.1 and 1.0);
\node[red!70!black] at (-2,-0.9) {$\mathcal F$};

\coordinate (Q)   at (-0.2,0.1);
\coordinate (TQ)  at (1.75,1.05);
\coordinate (PTQ) at (0.9,-0.5);

\node[fill=black, circle, inner sep=1.3pt,
      label={[xshift=-3pt,yshift=-5pt]above left:{$Q$}}] at (Q) {};
\node[fill=black, circle, inner sep=1.3pt,
      label={[xshift=3pt,yshift=2pt]above right:{$\mathcal T Q$}}] at (TQ) {};
\node[fill=black, circle, inner sep=1.3pt,
      label={[xshift=1pt,yshift=-0pt]below right:{$\mathcal T_{\mathcal F}^{\nu_b}Q$}}] at (PTQ) {};

\draw[->, thick] (Q) -- (TQ)
    node[midway, above, xshift=-5pt] {$\mathcal T$};
\draw[->, thick] (TQ) -- (PTQ)
    node[midway, right, xshift=2pt, yshift=-4pt] {$\Pi_{\mathcal F}^{\nu_b}$};
\draw[->, thick, densely dashed] (Q) -- (PTQ)
    node[midway, below, xshift=-3pt] {$\mathcal T_{\mathcal F}^{\nu_b}$};

\end{tikzpicture}

\caption{Norm mismatch. The Bellman update \(\mathcal T Q\) may leave
\(\mathcal F\), so FQE projects it back using the offline-sampling norm
\(L^2(\nu_b)\).
This projection need not preserve the contraction property of policy evaluation
in the target-occupancy norm \(L^2(d_\beta)\).}
\label{fig::mismatch}
\end{figure}

\subsection{Verifying Condition~\ref{cond::weight-stability} for Linear and RKHS Classes}
\label{app:weight-stability}

\begin{remark}[Linear classes]
For linear classes, Condition~\ref{cond::weight-stability} follows from
Gram-matrix conditions weaker than a pointwise lower bound on
\(\widehat w_\beta/w_\beta\). Suppose that
\[
\mathcal F
=
\{f_\theta=\phi^\top\theta:\theta\in\Theta\},
\]
and define the estimated-weight and target-occupancy Gram matrices
\[
\widehat\Sigma_\phi
:=
\mathbb E_{\nu_b}\!\left[
\widehat w_\beta\phi\phi^\top
\right],
\qquad
\Sigma_{\beta,\phi}
:=
\mathbb E_{\nu_b}\!\left[
w_\beta\phi\phi^\top
\right]
=
\mathbb E_{d_\beta}\!\left[
\phi\phi^\top
\right].
\]
Then Condition~\ref{cond::weight-stability} holds whenever the restricted
generalized Rayleigh quotient below is positive, and one may take any
\(\lambda_{\mathcal F}\) satisfying
\[
0<\lambda_{\mathcal F}\le
\inf_{\substack{
v\in\Theta-\Theta\\
v^\top\Sigma_{\beta,\phi}v>0
}}
\frac{v^\top\widehat\Sigma_\phi v}
     {v^\top\Sigma_{\beta,\phi}v}.
\]
The restricted quotient is bounded below by the unrestricted quotient:
\[
\inf_{\substack{
v\in\Theta-\Theta\\
v^\top\Sigma_{\beta,\phi}v>0
}}
\frac{v^\top\widehat\Sigma_\phi v}
     {v^\top\Sigma_{\beta,\phi}v}
\ge
\inf_{\substack{
v\in\mathbb R^p\\
v^\top\Sigma_{\beta,\phi}v>0
}}
\frac{v^\top\widehat\Sigma_\phi v}
     {v^\top\Sigma_{\beta,\phi}v}.
\]
Thus, Condition~\ref{cond::weight-stability} follows from weighted Gram-matrix
comparability. In particular, if
\[
\lambda_{\min}(\widehat\Sigma_\phi)
\ge \underline\lambda>0,
\qquad
\lambda_{\max}(\Sigma_{\beta,\phi})
\le \overline\lambda<\infty,
\]
then one may take
\(\lambda_{\mathcal F}=\underline\lambda/\overline\lambda\).
If \(\Theta-\Theta\) contains a nonzero multiple of every direction in
\(\mathbb R^p\), as when \(\Theta\) has nonempty interior, the preceding
inequality over \(\mathbb R^p\) is an equality.
\end{remark}

\medskip
 
\begin{remark}[RKHS classes]
Let \(\mathcal H\) be an RKHS with feature map
\(K_x:=K(x,\cdot)\), and suppose that
\(\mathcal F\subseteq\mathcal H\). Define the weighted covariance operators
\[
\widehat{\mathcal C}
:=
\mathbb E_{\nu_b}\!\left[
\widehat w_\beta(X)\,K_X\otimes K_X
\right],
\qquad
\mathcal C_\beta
:=
\mathbb E_{d_\beta}\!\left[
K_X\otimes K_X
\right].
\]
For every \(h\in\mathcal H\),
\[
\mathbb E_{\nu_b}\!\left[\widehat w_\beta h^2\right]
=
\langle h,\widehat{\mathcal C}h\rangle_{\mathcal H},
\qquad
\mathbb E_{d_\beta}\!\left[h^2\right]
=
\langle h,\mathcal C_\beta h\rangle_{\mathcal H}.
\]
Therefore, Condition~\ref{cond::weight-stability} holds with
\(\lambda_{\mathcal F}=\underline c\) whenever
\[
\widehat{\mathcal C}
\succeq
\underline c\,\mathcal C_\beta
\]
on \(\operatorname{span}(\mathcal F-\mathcal F)\). Thus, for RKHS classes,
Condition~\ref{cond::weight-stability} reduces to covariance-operator comparability along the
directions represented by the fitted class.
\end{remark}

\subsection{Interpolation Bounds for Weight--Residual Interactions}
\label{appendix:supnorm}
\label{app:approxwts}

Define the inherent Bellman error in the target-occupancy norm and the relative
density-ratio error by
\[
\varepsilon_{\mathrm{Bell}}
:=
\sup_{Q\in\mathcal F}
\left\|
\mathcal TQ-\mathcal T_{\mathcal F,\beta}Q
\right\|_{2,d_\beta},
\qquad
\varepsilon_w
:=
\left\|
\frac{\widehat w_\beta}{w_\beta}-1
\right\|_{2,d_\beta}.
\]
Equivalently,
\[
\varepsilon_w^2
=
\mathbb E_{\nu_b}\!\left[
\frac{(\widehat w_\beta-w_\beta)^2}{w_\beta}
\right].
\]

Because \(\mathcal T_{\mathcal F,\beta}Q\) is the \(L^2(d_\beta)\) projection of
\(\mathcal TQ\) onto \(\mathcal F\), \(\varepsilon_{\mathrm{Bell}}\) measures
the inherent Bellman error of \(\mathcal F\) in the target-occupancy norm.
Condition~\ref{cond::bellman-target-interp} holds with \(\alpha>0\) when the
residual class
\[
\left\{
\mathcal TQ-\mathcal T_{\mathcal F,\beta}Q:Q\in\mathcal F
\right\}
\]
has suitable finite-dimensional or smoothness structure.

\paragraph{Linear function approximation.}
Consider a linear-feature class
\(\mathcal F=\{Q_\theta=\phi^\top\theta:\theta\in\Theta\}\), where
\(\phi=(\phi_1,\ldots,\phi_p)^\top\). Define
\[
\mathcal G_{\mathrm{res}}
:=
\operatorname{span}
\{r_0,\phi_1,\ldots,\phi_p,
P_\pi\phi_1,\ldots,P_\pi\phi_p\}.
\]
Let \(\psi=(\psi_1,\ldots,\psi_q)^\top\) be an
\(L^2(d_\beta)\)-basis of \(\mathcal G_{\mathrm{res}}\). Then
\(\Sigma_\psi:=\E_{d_\beta}[\psi\psi^\top]\) is positive definite. Set
\[
\kappa_{\mathrm{res}}
:=
\operatorname*{ess\,sup}_{x\sim d_\beta}
\psi(x)^\top\Sigma_\psi^{-1}\psi(x).
\]

\begin{proposition}[Linear residual interpolation]
\label{prop:linear-residual-interpolation}
If \(\kappa_{\mathrm{res}}<\infty\), then
Condition~\ref{cond::bellman-target-interp} holds with
\(\alpha=1\) and \(C_\infty=\sqrt{\kappa_{\mathrm{res}}}\).
\end{proposition}

The condition \(\kappa_{\mathrm{res}}<\infty\) follows from familiar feature
boundedness and Gram-matrix conditioning: if
\(\sup_x\|\psi(x)\|_2\le B_\psi\) and
\(\lambda_{\min}(\Sigma_\psi)\ge c_\psi>0\), then
\(C_\infty\le B_\psi/\sqrt{c_\psi}\). If each coordinate is bounded by
\(B\) and \(c_\psi\) is dimension-independent, then
\(C_\infty\le B\sqrt{q/c_\psi}=O(\sqrt p)\), since \(q\le2p+1\). For an RKHS
residual class that is uniformly bounded in RKHS norm, suppose the kernel and
the eigenfunctions of its integral operator on \(L^2(d_\beta)\) are uniformly
bounded. If the eigenvalues satisfy \(\lambda_j\lesssim j^{-2r}\) for
\(r>1/2\), one may take
\(\alpha=1-1/(2r)\)
\citep[Lemma~5.1]{mendelson2010regularization}. Related interpolation bounds
hold for signed convex hulls of suitable bases
\citep[Lemma~2]{van2014uniform} and for \(d\)-variate H\"older or Sobolev
residual classes of order \(s>d/2\) on regular bounded domains with a design
density bounded below, with \(\alpha=1-d/(2s)\)
\citep[Lemma~4]{bibaut2021sequential}.

Under Condition~\ref{cond::bellman-target-interp} with \(\alpha>0\), the
weight-error term obeys
\[
\sup_{Q\in\mathcal F}\varepsilon_{\mathrm{Bell},w}(Q)
\le
C_\infty
\varepsilon_w
\varepsilon_{\mathrm{Bell}}^\alpha.
\]
Thus, on the completeness branch, density-ratio error and inherent Bellman
error enter multiplicatively. The realizability branch of
Theorem~\ref{thm:completeness-realizability} instead controls the residual
locally near \(Q^\pi_{\mathcal F,\beta}\), producing its higher-order and
\(\varepsilon_{\mathrm{app}}\)-dependent terms. In both branches, the
small-error guarantee also requires control of \(\varepsilon_w\) and restricted
norm comparability.

\begin{corollary}[Linear-class refinement]
\label{cor:linear-endpoint-refinement}
Assume Conditions~\ref{cond::stationary}--\ref{cond::weight-stability} and the
conditions of Proposition~\ref{prop:linear-residual-interpolation}. Define
\(\varepsilon_{\mathrm{app}}\) and \(\varepsilon_{\mathrm{Bell}}\) as in
Theorem~\ref{thm:completeness-realizability}. There exists
\(C=C(M,\kappa_{\mathrm{res}})<\infty\) such that, if
\[
\gamma_w
:=
\bar\gamma_\beta\left(1+C\frac{\varepsilon_w}{\lambda_{\mathcal F}}\right)
<1,
\]
then, for every \(\tau\in(0,1)\) and \(K\in\mathbb N\), with probability at
least \(1-\tau\),
\[
\begin{aligned}
\|\widehat Q^{(K)}-Q^\pi\|_{2,d_\beta}
&\le
\gamma_w^K
\|\widehat Q^{(0)}-Q^\pi_{\mathcal F,\beta}\|_{2,d_\beta}
+\frac{\varepsilon_{\mathrm{app}}}{1-\bar\gamma_\beta}
\\
&\quad+
\frac{C}{1-\gamma_w}
\left[
\frac{\varepsilon_{\mathrm{stat}}(K,\tau)}
{\sqrt{\lambda_{\mathcal F}}}
+\frac{\varepsilon_w}{\lambda_{\mathcal F}}
\min\left\{
\varepsilon_{\mathrm{Bell}},
\frac{\varepsilon_{\mathrm{app}}}{1-\bar\gamma_\beta}
\right\}
\right].
\end{aligned}
\]
\end{corollary}

Thus, exact Bellman completeness or exact value-function realizability removes
the additive weight-error term. Under exact realizability, the direct
\(\varepsilon_w/\lambda_{\mathcal F}\) dependence is absorbed into
\(\gamma_w\), while
\(\lambda_{\mathcal F}\) continues to control norm comparability.

\subsection{Approximate Weights and Stationary Limits}

\paragraph{Contraction under approximate occupancy weights.}
The analysis extends beyond exact recovery of \(w_\beta\). Suppose
\(\overline w_\beta\) is a limiting weight and the corresponding projected
Bellman operator is contractive in \(L^2(\overline w_\beta\nu_b)\) with modulus
\(\overline\gamma<1\). Then the same inexact Picard argument applies with
\(\overline\gamma\) in place of \(\bar\gamma_\beta\) and with the projected fixed point
defined under the \(\overline w_\beta\nu_b\)-weighted norm. Thus, misspecified
weights can still improve stability if they move the regression geometry
toward one in which the Bellman operator is contractive.

A simple sufficient condition is multiplicative closeness to the occupancy
ratio. Suppose, for some \(\varepsilon\in[0,1)\),
\[
(1-\varepsilon)w_\beta(s,a)
\le
\overline w_\beta(s,a)
\le
(1+\varepsilon)w_\beta(s,a).
\]
Then \(L^2(\overline w_\beta\nu_b)\) and
\(L^2(w_\beta\nu_b)=L^2(d_\beta)\) are equivalent:
\[
\|f\|_{2,d_\beta}^2
\le
\frac{1}{1-\varepsilon}
\|f\|_{2,\overline w_\beta\nu_b}^2,
\qquad
\|f\|_{2,\overline w_\beta\nu_b}^2
\le
(1+\varepsilon)\|f\|_{2,d_\beta}^2.
\]
Because \(\mathcal T\) is a \(\bar\gamma_\beta\)-contraction in
\(L^2(d_\beta)\),
\[
\|\mathcal TQ-\mathcal TQ'\|_{2,\overline w_\beta\nu_b}
\le
\bar\gamma_\beta
\sqrt{\frac{1+\varepsilon}{1-\varepsilon}}\,
\|Q-Q'\|_{2,\overline w_\beta\nu_b}.
\]
Hence, whenever
\[
\bar\gamma_\beta\sqrt{\frac{1+\varepsilon}{1-\varepsilon}}<1,
\]
the Bellman operator, and therefore its projection onto a closed convex fitted
class, remains contractive in \(L^2(\overline w_\beta\nu_b)\). Small
multiplicative weight errors thus preserve contraction, with an inflated
contraction factor.

\paragraph{On-policy and ergodic limits.}
When \(\nu_b=\mu_\pi\), stationarity gives
\(\xi_{\pi,b}=d_\beta=\mu_\pi\) and hence \(w_\beta\equiv1\) for every
\(\beta\in(0,1)\). Occupancy-weighted FQE then reduces to ordinary FQE, and all
weight-related errors vanish. More generally, under ergodic on-policy
trajectory sampling, empirical
state-action frequencies converge to \(\mu_\pi\), so the population regression
norm approaches the target-policy stationary norm. Extending this observation
to finite-sample guarantees requires additional mixing and empirical-process
control.

\subsection{Implicit Stationary Weighting by Sampling}
\label{app:implicit-stationary-sampling}

Stationary weighting can also be achieved implicitly when the transition kernel
\(P\), and hence \(\mu_\pi\), is known. If we can sample
\((\widetilde S_i,\widetilde A_i,\widetilde S_i')\sim \mu_\pi\otimes P\)
directly, then the regression design is already matched to the target
stationary distribution. Because these samples do not contain rewards, let
\(\widehat r\) estimate the reward function from the offline data,
and run unweighted FQE on
\[
\{(\widetilde S_i,\widetilde A_i,
\widehat r(\widetilde S_i,\widetilde A_i),\widetilde S_i')\}_{i=1}^n .
\]
This is ordinary FQE using samples from the target-policy stationary
distribution, with \(w_\pi\equiv 1\) and a perturbed reward.

Define
\[
\widehat{\mathcal T}Q:=\widehat r+\gamma P_\pi Q,
\qquad
\widehat{\mathcal T}_{\mathcal F}:=\Pi_{\mathcal F}\widehat{\mathcal T},
\]
and let \(\widehat Q^\pi_{\mathcal F}\) be the fixed point of
\(\widehat{\mathcal T}_{\mathcal F}\). For the \(K\)-step iterate
\(\widehat Q^{(K)}\),
\[
\|\widehat Q^{(K)}-Q^\pi_{\mathcal F}\|_{2,\mu_\pi}
\le
\|\widehat Q^{(K)}-\widehat Q^\pi_{\mathcal F}\|_{2,\mu_\pi}
+
\|\widehat Q^\pi_{\mathcal F}-Q^\pi_{\mathcal F}\|_{2,\mu_\pi}.
\]
The first term is the usual finite-iteration and statistical error under
stationary sampling. The second term is the reward-estimation perturbation.

\begin{theorem}[Reward perturbation]
\label{theorem:reward-perturbation}
Suppose \(\mu_\pi\) is stationary for \(P_\pi\),
\(r_0,\widehat r\in L^2(\mu_\pi)\), and \(\mathcal F\) is a nonempty closed
convex subset of \(L^2(\mu_\pi)\). Then
\[
\|\widehat Q^\pi_{\mathcal F}-Q^\pi_{\mathcal F}\|_{2,\mu_\pi}
\le
\frac{1}{1-\gamma}
\|\widehat r-r_0\|_{2,\mu_\pi}.
\]
If, in addition, \(\mathcal F\) is a linear subspace of \(L^2(\mu_\pi)\), then
\[
\|\widehat Q^\pi_{\mathcal F}-Q^\pi_{\mathcal F}\|_{2,\mu_\pi}
\le
\frac{1}{1-\gamma}
\|\Pi_{\mathcal F}(\widehat r-r_0)\|_{2,\mu_\pi}.
\]
\end{theorem}

\section{Experiments}
\label{sec::appendix}

\subsection{Finite-State Experiments}
\label{app:toy_norm_mismatch}

\paragraph{Baird hub-and-spokes.}
We use the seven-state hub-and-spokes construction of
\citet{baird1995residual}. The solid action moves deterministically to the hub,
whereas the dashed action moves uniformly to one of the six spoke states. The
target policy always selects the solid action, and the behavior policy selects
it with probability $\kappa$. Decreasing $\kappa$ therefore increases the
mismatch between the behavior and target stationary distributions. We use the
standard seven-dimensional linear feature map, zero rewards,
ridge-regularized FQE, and exact stationary occupancy ratios. Because
$Q^\pi\equiv0$, the experiment isolates stability of the projected Bellman
iteration. Results are averaged across 100 independent datasets.

\paragraph{Sparse Garnet MDPs.}
We also consider Garnet MDPs with 100 states, four actions, and branching
factor five. Transition probabilities over the five successor states are
sampled from a Dirichlet distribution, and rewards are Gaussian. The target
policy is randomly generated, while the behavior policy is
$\varepsilon$-greedy with $\varepsilon=0.1$. The data-generating distribution
uses a reset mechanism that differs from the target-policy stationary
distribution. The five-dimensional linear feature class contains $Q^\pi$ but
is not Bellman complete. We compare unweighted FQE with FQE using the exact
target-to-offline occupancy ratio.

\subsection{Stochastic Continuing CartPole}
\label{app:cartpole-study}

\paragraph{Environment and policies.}
We use a stochastic continuing version of CartPole with state
$S=(x,\dot x,\theta,\dot\theta)$ and action $A\in\{0,1\}$. Transitions follow
the standard CartPole dynamics with Euler step $0.02$, followed by independent
Gaussian noise with standard deviation $0.001$ in each state coordinate. If
the proposed successor satisfies $|x|>2.4$ or $|\theta|>12$ degrees, it is
replaced by an independent draw from
$\operatorname{Unif}([-0.05,0.05]^4)$. The reset is part of the transition
kernel, so the process remains continuing.

The reward is
\[
r(s)
=
\left(2-\frac{|\theta|}{12\text{ degrees}}\right)
\left(2-\frac{|x|}{2.4}\right)-1.
\]
Both behavior and target policies are logistic. With
$c=(0.8,1.3,12,2)^\top$,
\[
\Pr(A=1\mid S=s)
=
\sigma(c^\top s/T),
\]
where $T=1$ for the behavior policy and $T=1.5$ for the target policy. We set
the Bellman and occupancy discounts to
\[
\gamma=\beta=0.98.
\]

\paragraph{Offline data.}
Each offline dataset contains $n=25{,}000$ transitions sampled from the
stationary behavior distribution after burn-in. The unweighted and
FORE-weighted estimators use the same rewards and transitions.

\paragraph{FORE and FQE.}
FORE uses 128 joint random Fourier features with bandwidth multiplier one and
ridge penalty $10^{-2}$. Estimated ratios are bounded to
$[10^{-4},50]$ and normalized to have empirical mean one. The ratio estimator
is fit without rewards.

The $Q$-class uses 64 random Fourier features with bandwidth $1.4$. Both
unweighted and FORE-weighted FQE use ridge penalty $10^{-5}$, analytic
expectations over the target action distribution, and the empirical projected
fixed point. The two estimators therefore differ only in the regression
weights: unweighted FQE assigns unit weight to each transition, whereas
FORE-FQE uses the normalized estimated occupancy ratio.

\paragraph{Evaluation.}
We compare the estimators across 50 independent offline datasets. Policy value
is the discounted return under the target policy from the native reset
distribution. A large independent Monte Carlo sample of target-policy
trajectories provides the reference value.

For dataset $j$, define the paired improvement in absolute value error by
\[
G_j
=
\left|
\widehat V_{j,\mathrm{unweighted}}-V^\pi
\right|
-
\left|
\widehat V_{j,\mathrm{FORE}}-V^\pi
\right|.
\]
We report the mean absolute error of each estimator, the mean paired
improvement $\overline G$, its 95\% paired-bootstrap confidence interval, and
the proportion of datasets for which FORE-FQE has smaller absolute error.

We also summarize the estimated occupancy weights, including their effective
sample size and tail behavior.

\section{Contraction Results and Proofs for Discounted-Occupancy FQE}

\label{appendix::stationary}

\subsection{Stationary-Limit Contraction}

\begin{lemma}[Bellman contraction for $Q$-functions under a stationary measure]
\label{lemma::Qcontraction}
If \(\mu_\pi\) is stationary for \(P_\pi\) and
\(r_0\in L^2(\mu_\pi)\), then, for all
\(Q_1,Q_2\in L^2(\mu_\pi)\),
\[
\|P_\pi(Q_1-Q_2)\|_{2,\mu_\pi}
\le
\|Q_1-Q_2\|_{2,\mu_\pi}.
\]
Consequently, the Bellman operator \(\mathcal T\) is a \(\gamma\)-contraction:
\[
\|\mathcal TQ_1-\mathcal TQ_2\|_{2,\mu_\pi}
\le
\gamma\|Q_1-Q_2\|_{2,\mu_\pi}.
\]
\end{lemma}

\begin{proof}
Let \(h:=Q_1-Q_2\).
Conditional Jensen's inequality gives
\[
\{P_\pi h(s,a)\}^2
\le
P_\pi(h^2)(s,a).
\]
Integrating with respect to \(\mu_\pi\) and using
\(\mu_\pi P_\pi=\mu_\pi\) yields
\[
\|P_\pi h\|_{2,\mu_\pi}^2
\le \int P_\pi(h^2)\,d\mu_\pi
=\int h^2\,d\mu_\pi
=\|h\|_{2,\mu_\pi}^2.
\]
Since \(\mathcal TQ_1-\mathcal TQ_2=\gamma P_\pi h\), it follows that
\[
\|\mathcal TQ_1-\mathcal TQ_2\|_{2,\mu_\pi}
\le \gamma\|Q_1-Q_2\|_{2,\mu_\pi},
\]
which proves both claims.
\end{proof}

\begin{lemma}[Contraction of the projected Bellman operator]
\label{lemma::projected_contraction}
Assume that \(\mu_\pi\) is stationary for \(P_\pi\),
\(r_0\in L^2(\mu_\pi)\), and \(\mathcal F\) is a nonempty closed convex
subset of \(L^2(\mu_\pi)\). Then
\(\mathcal{T}_{\mathcal{F}}=\Pi_{\mathcal F}\mathcal T\) is a
\(\gamma\)-contraction on \(L^2(\mu_\pi)\); that is, for all
\(Q_1,Q_2\in L^2(\mu_\pi)\),
\[
\|\mathcal{T}_{\mathcal{F}} Q_1 - \mathcal{T}_{\mathcal{F}} Q_2\|_{2,{\mu_\pi}}
\, \le\, \gamma\,\|Q_1 - Q_2\|_{2,{\mu_\pi}}.
\]
\end{lemma}
\begin{proof}
The metric projection onto a nonempty closed convex subset of a Hilbert space
is nonexpansive. Therefore, Lemma~\ref{lemma::Qcontraction} gives
\begin{align*}
\|\mathcal T_{\mathcal F}Q_1-\mathcal T_{\mathcal F}Q_2\|_{2,\mu_\pi}
&\le \|\mathcal TQ_1-\mathcal TQ_2\|_{2,\mu_\pi}\\
&\le \gamma\|Q_1-Q_2\|_{2,\mu_\pi},
\end{align*}
as claimed.
\end{proof}

\begin{proof}[Proof of Lemma~\ref{lem:bellman-contraction}]
Let \(h:=Q-Q'\). Conditional Jensen's inequality and
\eqref{eq:offline-restarted-occupancy} give
\[
\|P_\pi h\|_{2,d_\beta}^2
\le \int h^2\,d(d_\beta P_\pi)
\le \beta^{-1}\|h\|_{2,d_\beta}^2.
\]
Hence
\(\|\mathcal TQ-\mathcal TQ'\|_{2,d_\beta}
\le\bar\gamma_\beta\|Q-Q'\|_{2,d_\beta}\).
Condition~\ref{cond::convex} makes
\(\Pi_{\mathcal F,\beta}\) nonexpansive in \(L^2(d_\beta)\), so composing
with this projection proves the claim.
\end{proof}

\subsection{Proofs for FQE with Discounted-Occupancy Weights}
\label{app:discounted-occupancy-proofs}
\label{app:otherweights}

We first establish the contraction bound used throughout
Appendix~\ref{app:discounted-occupancy-fqe}.

\begin{lemma}[Bellman contraction under discounted-occupancy weighting]
\label{thm:discounted-occ-contraction}
Fix a policy \(\pi\) and \(\gamma\in(0,1)\). For \(\beta\in(0,1)\), let
\[
d_{\rho,\pi,\beta}
:=
(1-\beta)\sum_{t=0}^\infty \beta^t\,\rho_t,
\]
where \(\rho_t\) denotes the distribution of \((S_t,A_t)\) under initial
state distribution \(\rho\) and policy \(\pi\). Equivalently,
\[
d_{\rho,\pi,\beta}
=(1-\beta)(\rho\otimes\pi)+\beta\,d_{\rho,\pi,\beta}P_\pi.
\]
Here \(d_{\rho,\pi,\beta}P_\pi\) denotes its one-step push-forward under the
target-policy transition kernel. Then, for any
\(Q_1,Q_2\in L^2(d_{\rho,\pi,\beta})\),
\[
\|\mathcal TQ_1-\mathcal TQ_2\|_{2,d_{\rho,\pi,\beta}}
\le
\frac{\gamma}{\sqrt{\beta}}\,
\|Q_1-Q_2\|_{2,d_{\rho,\pi,\beta}}.
\]
If \(r_0\in L^2(d_{\rho,\pi,\beta})\), then \(\mathcal T\) maps this space
into itself and is a contraction whenever
\[
\frac{\gamma}{\sqrt{\beta}}<1,
\qquad\text{equivalently,}\qquad
\beta>\gamma^2.
\]
In particular, if \(\beta=\gamma\), then the contraction modulus is
\(\sqrt{\gamma}\). If a stationary state-action distribution \(\mu_\pi\)
exists and satisfies \(\mu_\pi P_\pi=\mu_\pi\), the same argument with
\(\mu_\pi\) gives contraction modulus \(\gamma\).
\end{lemma}

\begin{proof}
Write \(d:=d_{\rho,\pi,\beta}\) and \(\Delta:=Q_1-Q_2\). Conditional
Jensen's inequality gives
\[
\|P_\pi\Delta\|_{2,d}^2
\le
\int \Delta^2\,d(dP_\pi).
\]
The balance identity implies \(\beta dP_\pi\le d\). Hence
\[
\|P_\pi\Delta\|_{2,d}^2
\le
\beta^{-1}\|\Delta\|_{2,d}^2.
\]
Because \(\mathcal TQ_1-\mathcal TQ_2=\gamma P_\pi\Delta\), taking square
roots proves the stated bound. Its modulus is smaller than one exactly when
\(\beta>\gamma^2\).
\end{proof}

\begin{proof}[Proof of Theorem~\ref{thm:discounted-projected-contraction}]
Apply Lemma~\ref{thm:discounted-occ-contraction} with
\(\beta=\gamma'\) and
\(d_{\rho,\pi,\beta}=d_{\pi,\gamma'}\). It gives
\[
\|\mathcal TQ-\mathcal TQ'\|_{2,d_{\pi,\gamma'}}
\le
\bar\gamma_{\gamma'}
\|Q-Q'\|_{2,d_{\pi,\gamma'}}.
\]
Because \(\mathcal F\) is nonempty, closed, and convex in the Hilbert space
\(L^2(d_{\pi,\gamma'})\), its metric projection is uniquely defined and
nonexpansive. Hence
\begin{align*}
\|\mathcal T_{\mathcal F,\gamma'}Q
-\mathcal T_{\mathcal F,\gamma'}Q'\|_{2,d_{\pi,\gamma'}}
&\le
\|\mathcal TQ-\mathcal TQ'\|_{2,d_{\pi,\gamma'}}
\\
&\le
\bar\gamma_{\gamma'}
\|Q-Q'\|_{2,d_{\pi,\gamma'}}.
\end{align*}
The map \(\mathcal T_{\mathcal F,\gamma'}\) sends the closed subset
\(\mathcal F\) into itself. Since \(\bar\gamma_{\gamma'}<1\), Banach's
fixed-point theorem gives the unique fixed point and the geometric bound
\eqref{eq:discounted-population-recursion}.

It remains to prove \eqref{eq:discounted-approximation-bound}. Let
\(f^\star:=\Pi_{\mathcal F,\gamma'}Q^\pi\). Since
\(\mathcal TQ^\pi=Q^\pi\), nonexpansiveness gives
\begin{align*}
\|Q^\pi_{\mathcal F,\gamma'}-Q^\pi\|_{2,d_{\pi,\gamma'}}
&\le
\|Q^\pi_{\mathcal F,\gamma'}-f^\star\|_{2,d_{\pi,\gamma'}}
+\|f^\star-Q^\pi\|_{2,d_{\pi,\gamma'}}
\\
&=
\|\Pi_{\mathcal F,\gamma'}\mathcal T
Q^\pi_{\mathcal F,\gamma'}
-\Pi_{\mathcal F,\gamma'}\mathcal TQ^\pi
\|_{2,d_{\pi,\gamma'}}
+\|f^\star-Q^\pi\|_{2,d_{\pi,\gamma'}}
\\
&\le
\bar\gamma_{\gamma'}
\|Q^\pi_{\mathcal F,\gamma'}-Q^\pi\|_{2,d_{\pi,\gamma'}}
+\inf_{f\in\mathcal F}\|f-Q^\pi\|_{2,d_{\pi,\gamma'}}.
\end{align*}
Rearranging proves \eqref{eq:discounted-approximation-bound}. Setting
\(\gamma'=\gamma\) gives the first specialization. For the second, setting
\(\rho=\nu_b^S\) and \(\gamma'=\beta\) gives
\(\rho\otimes\pi=\xi_{\pi,b}\),
\(d_{\pi,\gamma'}=d_\beta\), and
\(\bar\gamma_{\gamma'}=\bar\gamma_\beta\).
\end{proof}

\begin{proof}[Proof of Lemma~\ref{lem:discounted-inexact-recursion}]
By the triangle inequality and
Theorem~\ref{thm:discounted-projected-contraction},
\begin{align*}
\|\widehat Q^{(k)}-Q^\pi_{\mathcal F,\gamma'}
\|_{2,d_{\pi,\gamma'}}
&\le
\|\widehat Q^{(k)}
-\mathcal T_{\mathcal F,\gamma'}\widehat Q^{(k-1)}
\|_{2,d_{\pi,\gamma'}}
\\
&\quad+
\|\mathcal T_{\mathcal F,\gamma'}\widehat Q^{(k-1)}
-\mathcal T_{\mathcal F,\gamma'}
Q^\pi_{\mathcal F,\gamma'}\|_{2,d_{\pi,\gamma'}}
\\
&\le
\eta_{k,\gamma'}+
\bar\gamma_{\gamma'}
\|\widehat Q^{(k-1)}-Q^\pi_{\mathcal F,\gamma'}
\|_{2,d_{\pi,\gamma'}}.
\end{align*}
Iterating this one-step inequality proves
\eqref{eq:discounted-inexact-recursion}.
\end{proof}

\begin{proof}[Proof of Theorem~\ref{thm:discounted-dr-identity}]
The balance equation \eqref{eq:discounted-balance}, with
\(\gamma'=\gamma\), implies that every integrable \(Q\) satisfies
\begin{equation}
\label{eq:discounted-balance-integrated}
(1-\gamma)\E_{\xi_0}[Q]
=
\E_{d_\gamma}[Q]-\gamma\E_{d_\gamma P_\pi}[Q].
\end{equation}
Consequently,
\begin{align*}
\E_{d_\gamma}[\mathcal TQ-Q]
&=
\E_{d_\gamma}[r_0]
+\gamma\E_{d_\gamma P_\pi}[Q]
-\E_{d_\gamma}[Q]
\\
&=
J_\gamma(\pi)-(1-\gamma)\E_{\xi_0}[Q].
\end{align*}
Since \(w_\gamma\nu_b=d_\gamma\) and
\(\E_{\xi_0}[Q]=\E_\rho[(\pi Q)(S_0)]\), the last display yields
\(\Psi_{\rm DR}(w_\gamma,Q)=J_\gamma(\pi)\). Subtracting this equality from
\(\Psi_{\rm DR}(w,Q)\) proves \eqref{eq:dr-exact-remainder}. If
\(w=w_\gamma\), the right-hand side vanishes. If \(Q=Q^\pi\), it vanishes
because \(\mathcal TQ^\pi-Q^\pi=0\).

For the first product bound, write the right-hand side of
\eqref{eq:dr-exact-remainder} as
\[
\E_{\nu_b}\!\left[
\frac{w-w_\gamma}{\sqrt{w_\gamma}}
\sqrt{w_\gamma}(\mathcal TQ-Q)
\right]
\]
and apply Cauchy--Schwarz. For the second, let
\(\Delta:=Q-Q^\pi\). Lemma~\ref{thm:discounted-occ-contraction} with
\(\beta=\gamma\) gives
\(\|P_\pi\Delta\|_{2,d_\gamma}
\le\gamma^{-1/2}\|\Delta\|_{2,d_\gamma}\). Therefore
\begin{align*}
\|\mathcal TQ-Q\|_{2,d_\gamma}
&=
\|\gamma P_\pi\Delta-\Delta\|_{2,d_\gamma}
\\
&\le
\gamma\|P_\pi\Delta\|_{2,d_\gamma}
+\|\Delta\|_{2,d_\gamma}
\\
&\le
(1+\sqrt\gamma)\|Q-Q^\pi\|_{2,d_\gamma}.
\end{align*}
Combining the two inequalities proves \eqref{eq:dr-product-bound}.
\end{proof}

\begin{proof}[Proof of Corollary~\ref{cor:discounted-drl-bound}]
Conditional on the auxiliary sample, the expectation of each summand in
\eqref{eq:dr-estimator} is
\(\Psi_{\rm DR}(\widehat w,\widehat Q)\). Hence
\begin{align*}
|\widehat J_{\rm DR}-J_\gamma(\pi)|
&\le
\left|
\frac1n\sum_{i=1}^n
\{\widehat\psi_i-E[\widehat\psi_i\mid
\widehat w,\widehat Q]\}
\right|
\\
&\quad+
|\Psi_{\rm DR}(\widehat w,\widehat Q)-J_\gamma(\pi)|.
\end{align*}
Conditional Chebyshev inequality bounds the first term by
\(\sigma_{\rm DR}/\sqrt{n\tau}\) with probability at least
\(1-\tau\), and Theorem~\ref{thm:discounted-dr-identity} bounds the second.
This proves \eqref{eq:drl-finite-sample-bound}. The sub-Gaussian refinement
follows from the standard two-sided Chernoff bound.

When \(\widehat Q=\widehat Q^{(K)}\), the triangle inequality,
Lemma~\ref{lem:discounted-inexact-recursion}, and
\eqref{eq:discounted-approximation-bound}, all with \(\gamma'=\gamma\), give
\begin{align*}
\|\widehat Q^{(K)}-Q^\pi\|_{2,d_\gamma}
&\le
\gamma^{K/2}
\|\widehat Q^{(0)}-Q^\pi_{\mathcal F,\gamma}
\|_{2,d_\gamma}
\\
&\quad+
\sum_{j=1}^K\gamma^{(K-j)/2}\eta_{j,\gamma}
+\frac{1}{1-\sqrt\gamma}
\inf_{f\in\mathcal F}\|f-Q^\pi\|_{2,d_\gamma}.
\end{align*}
Substitution into \eqref{eq:drl-finite-sample-bound} proves
\eqref{eq:drl-fqe-product-bound}.
\end{proof}

\begin{proof}[Proof of Corollary~\ref{cor:discounted-intercept}]
Write \(Q_{\mathcal F}:=Q^\pi_{\mathcal F,\gamma}\) and
\(H:=\mathcal TQ_{\mathcal F}-Q_{\mathcal F}\). Since
\(Q_{\mathcal F}=\Pi_{\mathcal F,\gamma}\mathcal TQ_{\mathcal F}\), the
function \(c\mapsto\E_{d_\gamma}[(H-c)^2]\) is minimized at \(c=0\) over a
neighborhood of zero. Its derivative at zero must therefore vanish, giving
\(\E_{d_\gamma}[H]=0\). Applying
\eqref{eq:discounted-balance-integrated} to \(Q_{\mathcal F}\) yields
\[
\E_{d_\gamma}[H]
=
J_\gamma(\pi)
-(1-\gamma)\E_\rho[(\pi Q_{\mathcal F})(S_0)].
\]
The left-hand side is zero, which proves
\eqref{eq:intercept-value-correctness}.
\end{proof}

\section{Technical Lemmas}
\label{app:technical-lemmas}

\subsection{Local Maximal Inequality}

Let \(O_1,\ldots,O_n\in\mathcal O\) be independent random variables. For any
function \(f:\mathcal O\to\mathbb R\), define
\begin{align}
    \|f\| := \sqrt{\frac{1}{n} \sum_{i=1}^n\mathbb{E}[f(O_i)^2]}.
\end{align}

For a star-shaped class of functions $\mathcal{F}$ and a radius $\delta \in (0,\infty)$,
define the localized Rademacher complexity
\[
\mathcal{R}_n(\mathcal{F}, \delta)
:=
\mathbb{E}\left[
\sup_{\substack{f \in \mathcal{F} \\ \|f\| \le \delta}}
\frac{1}{n} \sum_{i=1}^n \epsilon_i f(O_i)
\right],
\]
where $\epsilon_i$ are i.i.d.\ Rademacher random variables.

The following lemma restates a local maximal inequality from Lemma~11 of
\citet{foster2023orthogonal}; see also Lemma~11 of
\citet{van2025nonparametric}.

\begin{lemma}[Local maximal inequality]\label{lemma:loc_max_ineq}
Let $\mathcal{F}$ be a star-shaped class of functions satisfying
$\sup_{f\in\mathcal{F}} \|f\|_{\infty} \le M$.
Let $\delta = \delta_n \in (0,1)$ satisfy the critical radius condition
$\mathcal{R}_n(\mathcal{F},\delta) \le \delta^2$,
and suppose that, as $n\to\infty$,
\[
\frac{1}{\sqrt{n}}\sqrt{\log\log(1/\delta_n)} = o(\delta_n).
\]
Then there exists a constant $C>0$ such that, for all $\eta \in (0,1)$,
with probability at least $1 - \eta$, every $f \in \mathcal{F}$ satisfies
\begin{align*}
&\left|
\frac{1}{n}\sum_{i=1}^n
\bigl(f(O_i) - \mathbb{E}[f(O_i)]\bigr)
\right|
\le
C\Bigl(
    \delta^2
    + \delta\,\|f\|
\Bigr)
\\
&\qquad+\;
C\Bigl(
    \frac{\sqrt{\log(e/\eta)}\,\|f\|}{\sqrt{n}}
    + \frac{M\,\log(e/\eta)}{n}
\Bigr).
\end{align*}
\end{lemma}
\begin{proof}
Apply the cited one-sided inequality to the symmetric star-shaped class
\(\mathcal F_\pm:=\mathcal F\cup(-\mathcal F)\). This class has the same
envelope as \(\mathcal F\), and its localized Rademacher complexity is at most
twice that of \(\mathcal F\). Thus a universal multiple of \(\delta_n\) is a
critical radius for \(\mathcal F_\pm\); we absorb this multiple into the
constant below.
Lemma~11 in \citet{van2025nonparametric} gives a constant \(C>0\) such that,
for all \(u\ge1\), with probability at least \(1-e^{-u^2}\), every
\(f\in\mathcal F_\pm\) satisfies
\[
\frac{1}{n}\sum_{i=1}^n
\bigl(f(O_i) - \mathbb{E}[f(O_i)]\bigr)
\le
C\Bigl(
  \delta_n^2
  + \delta_n\,\|f\|
  + \frac{u\,\|f\|}{\sqrt{n}}
  + \frac{M\,u^2}{n}
\Bigr).
\]
Set \(u:=\sqrt{\log(e/\eta)}\), for which
\(e^{-u^2}=\eta/e\le\eta\). On the resulting event, apply the inequality to
both \(f\) and \(-f\). Then, with probability at least \(1-\eta\),
every \(f\in\mathcal F\) satisfies
\begin{align*}
&\left|
\frac{1}{n}\sum_{i=1}^n
\bigl(f(O_i) - \mathbb{E}[f(O_i)]\bigr)
\right|
\le
C\Bigl(
    \delta_n^2
    + \delta_n\,\|f\|
\Bigr)
\\
&\qquad+\;
C\Bigl(
    \frac{\sqrt{\log(e/\eta)}\,\|f\|}{\sqrt{n}}
    + \frac{M\,\log(e/\eta)}{n}
\Bigr).
\end{align*}
This is the claimed two-sided maximal inequality.
\end{proof}

The next lemma bounds the localized Rademacher complexity by the uniform
entropy integral in the form obtained from Theorem~2.1 of
\citet{van2011local}.

For any distribution \(\mathsf Q\) and any uniformly bounded function class
\(\mathcal F\), let
\(N(\varepsilon,\mathcal F,L^2(\mathsf Q))\) denote the
\(\varepsilon\)-covering number of \(\mathcal F\) under the
\(L^2(\mathsf Q)\) norm \citep{van1996weak}.
Define the uniform entropy integral of \(\mathcal{F}\) by
\begin{equation*}
\mathcal{J}(\delta, \mathcal{F})
:=
\int_{0}^{\delta}
\sup_{\mathsf Q}
\sqrt{\log N(\epsilon, \mathcal{F}, L^2(\mathsf Q))}\, d\epsilon ,
\end{equation*}
where the supremum is over all discrete probability distributions \(\mathsf Q\).

\begin{lemma}\label{lemma:local_rademacher_entropy}
Let \(\mathcal{F}\) be a star-shaped class of functions such that
\(\sup_{f\in\mathcal F}\|f\|_\infty \le M\). Then, for every \(\delta>0\),
\[
\mathcal{R}_n(\mathcal{F}, \delta)
\;\lesssim\;
\frac{1}{\sqrt n}\,\mathcal{J}(\delta,\mathcal{F})
\left(1+\frac{\mathcal{J}(\delta,\mathcal{F})}{\delta^2 \sqrt n}\right),
\]
where the implicit constant depends only on \(M\).
\end{lemma}
\begin{proof}[Proof of Lemma~\ref{lemma:local_rademacher_entropy}]
Rademacher symmetrization followed by Theorem~2.1 of
\citet{van2011local}, applied to the class localized by \(\|f\|\le\delta\)
and rescaled to have unit envelope, gives
\[
\mathcal R_n(\mathcal F,\delta)
\le
\frac{C_M}{\sqrt n}\mathcal J(\delta,\mathcal F)
\left\{1+
\frac{\mathcal J(\delta,\mathcal F)}{\delta^2\sqrt n}
\right\},
\]
where \(C_M\) depends only on \(M\). This is the claimed bound.
\end{proof}

\subsection{One-Step Error for an Inexact FQE Update}

Recall the occupancy density ratio
\(w_\beta=\mathrm d d_\beta/\mathrm d\nu_b\), and let
\(\widehat w_{\beta}\) denote an estimate of this ratio trained independently of
\(\mathcal D_n\), as required by Condition~\ref{cond::split}.

Let \(\mathcal F\) be a convex function class and let
\(\widehat Q^{\mathrm{init}}\in\mathcal F\) be an initial, possibly
data-dependent, estimate. Define the weighted population risk
\[
\widehat R_0(Q) = \E_{\nu_b}\!\left[
  w_{\beta}(S,A)\,
  \bigl\{
    R + \gamma(\pi \widehat Q^{\mathrm{init}})(S') - Q(S,A)
  \bigr\}^2  
\right].
\]

Following Algorithm~\ref{alg:fore-weighted-fqe}, define the empirical minimizer
\[
\widehat Q_n := \argmin_{Q \in \mathcal{F}} \widehat R_n(Q),
\]
where
\[
\widehat R_n(Q)
=
\sum_{i=1}^n
\widehat w_{\beta}(S_i,A_i)\,
\bigl\{
  R_i + \gamma(\pi \widehat Q^{\mathrm{init}})(S_i') - Q(S_i,A_i)
\bigr\}^2.
\]

\begin{lemma}[Projected Bellman update as an occupancy-weighted population-risk minimizer]
\label{lem:proj-pop-risk}
Assume Conditions~\ref{cond::stationary}, \ref{cond::convex},
\ref{cond::bounded}, and~\ref{cond::weight-coverage}. Then
\[
\mathcal{T}_{\mathcal{F},\beta}(\widehat Q^{\mathrm{init}})
= \argmin_{Q \in \mathcal{F}} \widehat R_0(Q).
\]
\end{lemma}

\begin{proof}
Because \(w_\beta=\mathrm d d_\beta/\mathrm d\nu_b\), the measures
\(w_\beta\nu_b\) and \(d_\beta\) coincide. By definition of the occupancy
projection, \(\mathcal T_{\mathcal F,\beta}(\widehat Q^{\mathrm{init}})\)
minimizes the \(L^2(w_\beta\nu_b)\) distance to
\(\mathcal T(\widehat Q^{\mathrm{init}})\) over \(\mathcal F\). Moreover,
\[
E\{R+\gamma(\pi\widehat Q^{\mathrm{init}})(S')\mid S,A\}
=\mathcal T(\widehat Q^{\mathrm{init}})(S,A).
\]
Expanding the square conditional on \((S,A)\) shows that replacing the
conditional Bellman target by the noisy target adds only
\(E_{\nu_b}[w_\beta\operatorname{Var}\{R+\gamma(\pi\widehat Q^{\mathrm{init}})(S')
\mid S,A\}]\), which does not depend on \(Q\). Hence
\(\mathcal T_{\mathcal F,\beta}(\widehat Q^{\mathrm{init}})\) also minimizes
\(\widehat R_0\) over \(\mathcal F\).
\end{proof}

Recall the critical radius
\begin{align*}
\delta_{n}
&:=
\inf\Bigl\{
\delta > 0 :
\mathcal{J}(\delta,  \mathcal F)
\le
\sqrt{n}\,\delta^2
\Bigr\}
\vee\sqrt{\frac{\log(en)}{n}}.
\end{align*}

\begin{lemma}
\label{lemma::empiricalprocess}
Let \(w\) be a state-action function with \(\|w\|_\infty\le B\), where
\(B\ge1\). Assume that \(R\) and \(\mathcal F\) are uniformly bounded by
\(M\), and that the critical radius \(\delta_n\) above is finite and satisfies
\(\delta_n<1\). Then there exists a constant
\(C=C(M)>0\) such that, for all \(\tau \in (0,1)\), with probability at least
\(1-\tau\), simultaneously for all \(Q_1,Q_2\in\mathcal F\) and
\(V_1\in\mathcal V_{\mathcal F}:=\{\pi Q:Q\in\mathcal F\}\),
\begin{align*}
&\left|(P_n-P_0)\big[
   w\,
   (Q_1-Q_2)\,
   \{R+\gamma V_1-Q_2\}
 \big]\right|
\\
& \le
C\Bigl(
    B(\delta_n)^2
    + \delta_n\,\|w(Q_1-Q_2)\|
\Bigr)
\\
&\qquad+\;
C\Bigl(
    \frac{\sqrt{\log(e/\tau)}\,\|w(Q_1-Q_2)\|}{\sqrt{n}}
    + \frac{B\log(e/\tau)}{n}
\Bigr),
\end{align*}
where $\|\cdot\|$ denotes the offline-sampling norm \(L^2(P_0)\).
\end{lemma}
\begin{proof}
Fix a bounded weight function \(w\) with \(\|w\|_\infty \le B\). If \(w\) is
estimated on an independent sample, condition on that sample throughout. Let
\[
\mathcal V_{\mathcal F}
:=
\{\,\pi Q : Q\in\mathcal F\,\},
\qquad
(\pi Q)(s):=\sum_{a\in\mathcal A}\pi(a\mid s)Q(s,a).
\]
Define the function class
\begin{align*}
\mathcal{G}_w
:=
\Bigl\{
&(s,a,r,s') \mapsto
w(s,a)\,
\bigl(Q_1(s,a) - Q_2(s,a)\bigr)
\\[-0.4em]
&\times
\bigl(r + \gamma V_1(s') - Q_2(s,a)\bigr)
:\;
Q_1, Q_2 \in \mathcal{F},\;
V_1 \in \mathcal V_{\mathcal F}
\Bigr\}.
\end{align*}
By boundedness, each $g\in\mathcal G_w$ satisfies
\[
|g(s,a,r,s')|
\le
C(M)\,\bigl|w(s,a)\{Q_1(s,a)-Q_2(s,a)\}\bigr|,
\]
and hence
\begin{equation}
\label{eq:norm-g-vs-wdiff-fqe}
\|g\|_{L^2(P_0)}
\le
C(M)\,\|w(Q_1-Q_2)\|_{L^2(P_0)}.
\end{equation}

\paragraph{Entropy control.}
For any discrete distribution \(\mathsf Q_S\) on \(\mathcal S\), let
\(\bar{\mathsf Q}\) be its target-policy lift to the state--action space:
\[
\bar{\mathsf Q}(s,a):=\mathsf Q_S(s)\pi(a\mid s).
\]
For any \(Q,\widetilde Q\in\mathcal F\), conditional Jensen's inequality gives
\[
\bigl|\pi(Q-\widetilde Q)(s)\bigr|^2
\le
\sum_{a\in\mathcal A}\pi(a\mid s)
\{Q(s,a)-\widetilde Q(s,a)\}^2.
\]
Therefore
\[
\|\pi Q-\pi\widetilde Q\|_{L^2(\mathsf Q_S)}
\le
\|Q-\widetilde Q\|_{L^2(\bar{\mathsf Q})}.
\]
It follows that, for every \(\varepsilon>0\),
\[
N(\varepsilon,\mathcal V_{\mathcal F},L^2(\mathsf Q_S))
\le
N(\varepsilon,\mathcal F,L^2(\bar{\mathsf Q})).
\]
Since the entropy supremum for \(\mathcal F\) ranges over all discrete
state--action distributions, the uniform entropy of
\(\mathcal V_{\mathcal F}\) is bounded by that of \(\mathcal F\).

Let \(\mathcal G_{w/B}:=\{g/B:g\in\mathcal G_w\}\). The map
\((Q_1,Q_2,V_1)\mapsto (w/B)(Q_1-Q_2)(r+\gamma V_1-Q_2)\) is a
Lipschitz pointwise transformation of uniformly bounded functions
with multiplier envelope one. Standard permanence properties of entropy under
Lipschitz maps and bounded multipliers \citep{van1996weak} imply that there is
a constant \(C_1=C_1(M)\) such that, for all \(\delta>0\),
\[
\mathcal{J}(\delta,\mathcal G_{w/B})
\le
C_1\,\mathcal{J}(C_1\delta,\mathcal F).
\]
Thus \(\mathcal G_{w/B}\) has the same critical radius as \(\mathcal F\), up to
constants absorbed into \(C\).

Let
\[
\mathcal H_{w/B}
:=
\{\,t g:g\in\mathcal G_{w/B},\ t\in[-1,1]\,\}
\]
be the symmetric star-shaped hull, and let \(M_G=C(M)\) be an envelope for
\(\mathcal G_{w/B}\). The covering bound for star-shaped hulls gives, for all
\(\delta>0\),
\[
\mathcal J(\delta,\mathcal H_{w/B})
\;\lesssim\;
\mathcal J(C_1\delta,\mathcal G_{w/B})
+
\delta\sqrt{\log\!\left(1+\frac{M_G}{\delta}\right)}.
\]
The additional term is at most a constant multiple of \(\sqrt n\,\delta^2\)
whenever \(\delta\gtrsim\sqrt{\log(eM_Gn)/n}\). Since \(M_G\) is structural and
the definition of \(\delta_n\) includes the parametric floor, \(\delta_n\) is a
critical radius for \(\mathcal H_{w/B}\), up to constants depending only on
\(M\). The same floor also makes the side condition in
Lemma~\ref{lemma:loc_max_ineq} automatic: for all sufficiently large \(n\)
with \(\delta_n<1\),
\[
\frac{n^{-1/2}\sqrt{\log\log(1/\delta_n)}}{\delta_n}
\le
\sqrt{
\frac{\log\log\{\sqrt{n/\log(en)}\}}
{\log(en)}
}
=o(1).
\]
For the finitely many remaining sample sizes, boundedness gives the same
conclusion after enlarging the constant.
Lemmas~\ref{lemma:local_rademacher_entropy} and~\ref{lemma:loc_max_ineq}
applied to \(\mathcal H_{w/B}\) give that, with probability at least
\(1-\tau\), every \(h\in\mathcal G_{w/B}\) satisfies
\begin{align*}
|(P_n-P_0)h|
&\le
C\Bigl(
    (\delta_n)^2
    + \delta_n\,\|h\|_{L^2(P_0)}
\Bigr)
\\
&\qquad+\;
C\Bigl(
    \frac{\sqrt{\log(e/\tau)}\,\|h\|_{L^2(P_0)}}{\sqrt{n}}
    + \frac{\log(e/\tau)}{n}
\Bigr).
\end{align*}
Multiplying this display by \(B\) gives that every \(g\in\mathcal G_w\)
satisfies
\begin{align*}
|(P_n-P_0)g|
&\le
C\Bigl(
    B(\delta_n)^2
    + \delta_n\,\|g\|_{L^2(P_0)}
\Bigr)
\\
&\qquad+\;
C\Bigl(
    \frac{\sqrt{\log(e/\tau)}\,\|g\|_{L^2(P_0)}}{\sqrt{n}}
    + \frac{B\,\log(e/\tau)}{n}
\Bigr).
\end{align*}
Using \eqref{eq:norm-g-vs-wdiff-fqe} yields the claimed uniform bound.
Because the event is simultaneous over \(Q_1,Q_2\in\mathcal F\) and
\(V_1\in\mathcal V_{\mathcal F}\), it can later be specialized to fitted,
data-dependent choices such as
\(Q_1=\mathcal T_{\mathcal F,\beta}(\widehat Q^{\mathrm{init}})\),
\(Q_2=\widehat Q_n\), and \(V_1=\pi\widehat Q^{\mathrm{init}}\).
\end{proof}

\begin{proof}[Proof of Lemma~\ref{lemma::errorperiter}]
Fix an iteration \(k\), and write
\(\widehat Q^{\mathrm{init}}=\widehat Q^{(k)}\) and
\(\widehat Q_n=\widehat Q^{(k+1)}\). Under occupancy weighting,
\(w_\beta=\mathrm d d_\beta/\mathrm d\nu_b\), so
\(w_\beta\nu_b=d_\beta\). For brevity, define
\[
\|f\|_{2,\widehat w_{\beta}}^2 := P_0\big[\widehat w_{\beta} f^2\big].
\]
We first use the empirical first-order condition, then lower bound its
population counterpart and control the remaining empirical-process term.

By first-order optimality of
\(\widehat Q_n = \argmin_{Q \in \mathcal{F}} \widehat R_n(Q)\) and convexity of
\(\mathcal{F}\), we have, for all \(Q \in \mathcal{F}\),
\begin{align*}
\frac{1}{n}\sum_{i=1}^n
&\;\widehat w_{\beta}(S_i,A_i)\,
\bigl\{Q(S_i,A_i)-\widehat Q_n(S_i,A_i)\bigr\}
\\
&\qquad\times
\bigl\{R_i + \gamma\,(\pi \widehat Q^{\mathrm{init}})(S_i') - \widehat Q_n(S_i,A_i)\bigr\}
\le 0.
\end{align*}
Since
\(\mathcal{T}_{\mathcal{F},\beta}(\widehat Q^{\mathrm{init}}) \in \mathcal{F}\),
taking \(Q = \mathcal{T}_{\mathcal{F},\beta}(\widehat Q^{\mathrm{init}})\) gives
\begin{align*}
\frac{1}{n}\sum_{i=1}^n
&\;\widehat w_{\beta}(S_i,A_i)\,
\bigl\{\mathcal{T}_{\mathcal{F},\beta}(\widehat Q^{\mathrm{init}})(S_i,A_i)
- \widehat Q_n(S_i,A_i)\bigr\}
\\[-0.25em]
&\qquad\qquad\times
\bigl\{R_i + \gamma\,(\pi \widehat Q^{\mathrm{init}})(S_i') - \widehat Q_n(S_i,A_i)\bigr\}
\le 0.
\end{align*}
Adding and subtracting the population expectation and rearranging,
\begin{equation}
\label{eqn::firstbasic}
\begin{aligned}
&P_0\big[
   \widehat w_{\beta}\,
   (\mathcal{T}_{\mathcal{F},\beta}(\widehat Q^{\mathrm{init}}) - \widehat Q_n)\,
   \{R + \gamma(\pi \widehat Q^{\mathrm{init}}) - \widehat Q_n\}
\big]
\\
&\le
(P_0 - P_n)\big[
   \widehat w_{\beta}\,
   (\mathcal{T}_{\mathcal{F},\beta}(\widehat Q^{\mathrm{init}}) - \widehat Q_n)\,
   \{R + \gamma(\pi \widehat Q^{\mathrm{init}}) - \widehat Q_n\}
\big].
\end{aligned}
\end{equation}
In the displays below, \(R\), \(\widehat Q_n\), and
\(\pi\widehat Q^{\mathrm{init}}\) denote their coordinate evaluations at
\((S,A,R,S')\); in particular, the Bellman target uses
\((\pi\widehat Q^{\mathrm{init}})(S')\).

Set
\[
\Delta:=\mathcal T_{\mathcal F,\beta}(\widehat Q^{\mathrm{init}})-\widehat Q_n,
\qquad
b:=\mathcal T(\widehat Q^{\mathrm{init}})
-\mathcal T_{\mathcal F,\beta}(\widehat Q^{\mathrm{init}}).
\]
Taking conditional expectations given \((S,A)\) gives
\[
P_0\!\left[
   \widehat w_\beta \Delta
   \{R+\gamma(\pi\widehat Q^{\mathrm{init}})-\widehat Q_n\}
\right]
=
P_0[\widehat w_\beta \Delta(\Delta+b)].
\]
Because \(\Delta\in\mathcal H_{\mathcal F}\),
Condition~\ref{cond::weight-stability} gives
\[
\|\Delta\|_{2,d_\beta}
\le
\lambda_{\mathcal F}^{-1/2}\|\Delta\|_{2,\widehat w_\beta}.
\]
Moreover, \(w_\beta\nu_b=d_\beta\), and the variational inequality for projection
onto a convex set gives \(P_0[w_\beta \Delta b]\ge0\): the projection point is
\(\mathcal T_{\mathcal F,\beta}(\widehat Q^{\mathrm{init}})\), the comparison point is
\(\widehat Q_n\in\mathcal F\), and \(b\) is the residual from the unprojected
Bellman target. Since \(w_\beta\ge\underline\kappa_\beta>0\),
\(\widehat w_\beta-w_\beta
=w_\beta(\widehat w_\beta/w_\beta-1)\) \(\nu_b\)-almost surely. Therefore, by
Cauchy--Schwarz and the definition of
\(\varepsilon_{\mathrm{Bell},w}(\widehat Q^{\mathrm{init}})\),
\[
P_0[\widehat w_\beta \Delta b]
=
P_0[w_\beta \Delta b]
+
P_0\!\left[
  w_\beta
  \left(\frac{\widehat w_\beta}{w_\beta}-1\right)
  \Delta b
\right]
\ge
-
\frac{\varepsilon_{\mathrm{Bell},w}(\widehat Q^{\mathrm{init}})}
     {\sqrt{\lambda_{\mathcal F}}}
\|\Delta\|_{2,\widehat w_\beta}.
\]
Combining these inequalities yields
\[
P_0\!\left[
   \widehat w_\beta \Delta
   \{R+\gamma(\pi\widehat Q^{\mathrm{init}})-\widehat Q_n\}
\right]
\ge
\|\Delta\|_{2,\widehat w_\beta}^2
-
\frac{\varepsilon_{\mathrm{Bell},w}(\widehat Q^{\mathrm{init}})}
     {\sqrt{\lambda_{\mathcal F}}}
\|\Delta\|_{2,\widehat w_\beta}.
\]
Applying this lower bound to \eqref{eqn::firstbasic} and rearranging,
\begin{equation}
\label{eqn::secondbasic}
\begin{aligned}
&\|\Delta\|_{2,\widehat w_{\beta}}^2
\\
&\le
(P_0 - P_n)\big[
   \widehat w_{\beta}\,
   \Delta\,
   \{R + \gamma(\pi \widehat Q^{\mathrm{init}}) - \widehat Q_n\}
\big]
\\
&\quad+
\frac{\varepsilon_{\mathrm{Bell},w}(\widehat Q^{\mathrm{init}})}
     {\sqrt{\lambda_{\mathcal F}}}
\|\Delta\|_{2,\widehat w_{\beta}}.
\end{aligned}
\end{equation}

We now invoke Condition~\ref{cond::split}. Conditional on the independent
sample used to estimate the weights, \(\widehat w_\beta\) is a fixed bounded
multiplier. Lemma~\ref{lemma::empiricalprocess} applies uniformly over the
regression and Bellman-target classes, so it may be specialized to the
data-dependent fitted functions
\(\mathcal T_{\mathcal F,\beta}(\widehat Q^{\mathrm{init}})\), \(\widehat Q_n\), and
\(\pi\widehat Q^{\mathrm{init}}\) even though the FQE regressions reuse the same
Bellman-regression sample. Condition~\ref{cond::weight-coverage} gives
\[
\|\widehat w_\beta\Delta\|_{L^2(P_0)}^2
\le
\kappa_{\mathrm{cov}}
\|\Delta\|_{2,\widehat w_\beta}^2.
\]
Applying Lemma~\ref{lemma::empiricalprocess} with
\(w=\widehat w_\beta\), and then integrating over the independent
weight-estimation sample, gives, for \(C=C(M)>0\), with probability at least
\(1-\tau\),
\begin{align*}
&\left|(P_0 - P_n)\big[
   \widehat w_{\beta}\,
   \Delta\,
   \{R + \gamma(\pi \widehat Q^{\mathrm{init}}) - \widehat Q_n\}
\big]\right|
\\
&\le
C\kappa_{\mathrm{cov}}
\left(
    (\delta_n)^2
    +\frac{\log(e/\tau)}{n}
\right)
\\
&\quad+
C\sqrt{\kappa_{\mathrm{cov}}}
\left(
    \delta_n+\sqrt{\frac{\log(e/\tau)}{n}}
\right)
\|\Delta\|_{2,\widehat w_{\beta}}.
\end{align*}

Combining this high-probability bound with \eqref{eqn::secondbasic}, and
writing
\[
y:=\|\Delta\|_{2,\widehat w_\beta},
\qquad
a:=\delta_n+\sqrt{\log(e/\tau)/n},
\qquad
\omega:=\varepsilon_{\mathrm{Bell},w}(\widehat Q^{\mathrm{init}}),
\]
gives
\[
y^2
\le
C\kappa_{\mathrm{cov}}a^2
+
\left(
C\sqrt{\kappa_{\mathrm{cov}}}\,a
+
\frac{\omega}{\sqrt{\lambda_{\mathcal F}}}
\right)y .
\]
Solving this scalar inequality yields
\[
y
\le
C\left\{
\sqrt{\kappa_{\mathrm{cov}}}\,a
+
\frac{\omega}{\sqrt{\lambda_{\mathcal F}}}
\right\}.
\]
Condition~\ref{cond::weight-stability} and
\(\sqrt{\kappa_{\mathrm{cov}}}a=\varepsilon_{\mathrm{stat}}(1,\tau)\)
therefore give
\[
\|\mathcal T_{\mathcal F,\beta}(\widehat Q^{\mathrm{init}})-\widehat Q_n\|_{2,d_\beta}
\le
C\left\{
\frac{\varepsilon_{\mathrm{stat}}(1,\tau)}
     {\sqrt{\lambda_{\mathcal F}}}
+
\frac{\varepsilon_{\mathrm{Bell},w}(\widehat Q^{\mathrm{init}})}
     {\lambda_{\mathcal F}}
\right\},
\]
which is the bound in Lemma~\ref{lemma::errorperiter}.
\end{proof}

\section{Proofs of Main Results}
\label{app:main-proofs}

\begin{theorem}[Function approximation error]
\label{lem:proj-fixed-point-error}
Under Conditions~\ref{cond::stationary}--\ref{cond::convex},
\[
\|Q^\pi_{\mathcal F,\beta}-Q^\pi\|_{2,d_\beta}
\le
\frac{1}{1-\bar\gamma_\beta}
\inf_{f\in\mathcal F}\|f-Q^\pi\|_{2,d_\beta}.
\]
\end{theorem}

\begin{proof}[Proof of Theorem~\ref{lem:proj-fixed-point-error}]
Let \(f^\star:=\Pi_{\mathcal F,\beta}Q^\pi\). The fixed-point identities,
\(\mathcal TQ^\pi=Q^\pi\), and Lemma~\ref{lem:bellman-contraction} give
\begin{align*}
\|Q^\pi_{\mathcal F,\beta}-Q^\pi\|_{2,d_\beta}
&\le
\|Q^\pi_{\mathcal F,\beta}-f^\star\|_{2,d_\beta}
+\|f^\star-Q^\pi\|_{2,d_\beta}
\\
&=
\|\mathcal T_{\mathcal F,\beta}Q^\pi_{\mathcal F,\beta}
-\mathcal T_{\mathcal F,\beta}Q^\pi\|_{2,d_\beta}
+\|f^\star-Q^\pi\|_{2,d_\beta}
\\
&\le
\bar\gamma_\beta
\|Q^\pi_{\mathcal F,\beta}-Q^\pi\|_{2,d_\beta}
+\inf_{f\in\mathcal F}\|f-Q^\pi\|_{2,d_\beta}.
\end{align*}
Rearranging proves the stated bound.
\end{proof}

\subsection{Proofs of the Completeness and Realizability Refinements}
\label{app:proof-completeness-realizability}

\begin{proof}[Proof of Proposition~\ref{prop:linear-residual-interpolation}]
For \(Q_\theta=\phi^\top\theta\), write
\(\mathcal T_{\mathcal F,\beta}Q_\theta=\phi^\top\overline\theta\) for some
\(\overline\theta\in\Theta\). Then
\[
\mathcal TQ_\theta-\mathcal T_{\mathcal F,\beta}Q_\theta
=
r_0+\gamma(P_\pi\phi)^\top\theta-\phi^\top\overline\theta
\in\mathcal G_{\mathrm{res}}.
\]
For any \(g=a^\top\psi\in\mathcal G_{\mathrm{res}}\), Cauchy--Schwarz gives,
for \(d_\beta\)-almost every \(x\),
\[
|g(x)|
\le
\{a^\top\Sigma_\psi a\}^{1/2}
\{\psi(x)^\top\Sigma_\psi^{-1}\psi(x)\}^{1/2}
\le
\sqrt{\kappa_{\mathrm{res}}}\,\|g\|_{2,d_\beta}.
\]
Applying this inequality to each Bellman projection residual proves
Condition~\ref{cond::bellman-target-interp} with \(\alpha=1\) and
\(C_\infty=\sqrt{\kappa_{\mathrm{res}}}\).
\end{proof}

\begin{proof}[Proof of Theorem~\ref{thm:completeness-realizability}]
Let
\[
e_k
:=
\|\widehat Q^{(k)}-Q^\pi_{\mathcal F,\beta}\|_{2,d_\beta},
\qquad
B(Q)
:=
\mathcal TQ-\mathcal T_{\mathcal F,\beta}Q.
\]
Throughout the proof, \(C=C(M,\alpha,\beta,\gamma)\) may change from line to
line and is independent of \(C_\infty\), \(\varepsilon_w\), and
\(\lambda_{\mathcal F}\).
We first bound the projection residual using value-function approximation. Let
\(Q_0:=\Pi_{\mathcal F,\beta}Q^\pi\), so
\(\|Q_0-Q^\pi\|_{2,d_\beta}=\varepsilon_{\mathrm{app}}\). Since
\(Q^\pi_{\mathcal F,\beta}\) is the fixed point of \(\mathcal T_{\mathcal F,\beta}\),
the definition of metric projection gives
\begin{align*}
\|B(Q^\pi_{\mathcal F,\beta})\|_{2,d_\beta}
&=
\operatorname{dist}_{2,d_\beta}
\bigl(\mathcal TQ^\pi_{\mathcal F,\beta},\mathcal F\bigr)
\\
&\le
\|\mathcal TQ^\pi_{\mathcal F,\beta}-Q_0\|_{2,d_\beta}
\\
&\le
\bar\gamma_\beta\|Q^\pi_{\mathcal F,\beta}-Q^\pi\|_{2,d_\beta}
+\varepsilon_{\mathrm{app}}
\le
\frac{\varepsilon_{\mathrm{app}}}{1-\bar\gamma_\beta},
\end{align*}
where the last inequality uses Theorem~\ref{lem:proj-fixed-point-error}.
Because distance to a closed set is nonexpansive and \(\mathcal T\) is a
\(\bar\gamma_\beta\)-contraction in \(L^2(d_\beta)\), it follows that
\begin{equation}
\label{eq:bellman-residual-realizability}
\|B(\widehat Q^{(k)})\|_{2,d_\beta}
\le
\min\left\{
\varepsilon_{\mathrm{Bell}},
\bar\gamma_\beta e_k+\frac{\varepsilon_{\mathrm{app}}}{1-\bar\gamma_\beta}
\right\}.
\end{equation}

Condition~\ref{cond::bellman-target-interp}, Cauchy--Schwarz, and
\eqref{eq:bellman-residual-realizability} imply
\begin{align}
\label{eq:pointwise-weight-residual-refinement}
\varepsilon_{\mathrm{Bell},w}(\widehat Q^{(k)})
&\le
\varepsilon_w
\|B(\widehat Q^{(k)})\|_\infty
\nonumber\\
&\le
C_\infty\varepsilon_w
\min\left\{
\varepsilon_{\mathrm{Bell}}^\alpha,
\left(\bar\gamma_\beta e_k+
\frac{\varepsilon_{\mathrm{app}}}{1-\bar\gamma_\beta}\right)^\alpha
\right\}.
\end{align}
Apply Lemma~\ref{lemma::errorperiter} at confidence level \(\tau/K\), take a
union bound over the \(K\) regressions, and combine the resulting pointwise
one-step bounds with Lemma~\ref{lem:bellman-contraction}. On an event of
probability at least \(1-\tau\), for every \(0\le k<K\),
\begin{equation}
\label{eq:combined-structure-recursion}
e_{k+1}
\le
\bar\gamma_\beta e_k
+C\frac{\varepsilon_{\mathrm{stat}}(K,\tau)}
{\sqrt{\lambda_{\mathcal F}}}
+C\frac{C_\infty\varepsilon_w}{\lambda_{\mathcal F}}
\min\left\{
\varepsilon_{\mathrm{Bell}}^\alpha,
\left(\bar\gamma_\beta e_k+
\frac{\varepsilon_{\mathrm{app}}}{1-\bar\gamma_\beta}\right)^\alpha
\right\}.
\end{equation}

Using the completeness branch in \eqref{eq:combined-structure-recursion} and
\(\bar\gamma_\beta\le\gamma_{\beta,\star}\) gives
\[
e_{k+1}
\le
\gamma_{\beta,\star} e_k
+C\frac{\varepsilon_{\mathrm{stat}}(K,\tau)}
{\sqrt{\lambda_{\mathcal F}}}
+C\frac{C_\infty\varepsilon_w}{\lambda_{\mathcal F}}
\varepsilon_{\mathrm{Bell}}^\alpha.
\]
For the realizability branch, suppose first that \(\alpha\in(0,1)\). By
subadditivity of \(t\mapsto t^\alpha\) and Young's inequality with coefficient
\(\gamma_{\beta,\star}-\bar\gamma_\beta>0\),
\[
C\frac{C_\infty\varepsilon_w}{\lambda_{\mathcal F}}
\left(\bar\gamma_\beta e_k+
\frac{\varepsilon_{\mathrm{app}}}{1-\bar\gamma_\beta}\right)^\alpha
\le
(\gamma_{\beta,\star}-\bar\gamma_\beta)e_k
+C\bigl\{
\left(\frac{C_\infty\varepsilon_w}{\lambda_{\mathcal F}}\right)^{1/(1-\alpha)}
+\frac{C_\infty\varepsilon_w}{\lambda_{\mathcal F}}
\left(\frac{\varepsilon_{\mathrm{app}}}{1-\bar\gamma_\beta}\right)^\alpha
\bigr\}.
\]
When \(\alpha=0\), the same bound follows directly from the convention
\(x^0=1\), after enlarging \(C\).
Therefore both branches imply
\[
e_{k+1}
\le
\gamma_{\beta,\star} e_k
+C\frac{\varepsilon_{\mathrm{stat}}(K,\tau)}
{\sqrt{\lambda_{\mathcal F}}}
+C\min\!\left\{
\frac{C_\infty\varepsilon_w}{\lambda_{\mathcal F}}
\varepsilon_{\mathrm{Bell}}^\alpha,
\left(\frac{C_\infty\varepsilon_w}{\lambda_{\mathcal F}}\right)^{1/(1-\alpha)}
+\frac{C_\infty\varepsilon_w}{\lambda_{\mathcal F}}
\left(\frac{\varepsilon_{\mathrm{app}}}{1-\bar\gamma_\beta}\right)^\alpha
\right\}.
\]
Unrolling this recursion and using
\(\sum_{j=0}^{K-1}\gamma_{\beta,\star}^j
\le(1-\gamma_{\beta,\star})^{-1}\) bounds
\(e_K\). Finally, Theorem~\ref{lem:proj-fixed-point-error} gives
\[
\|\widehat Q^{(K)}-Q^\pi\|_{2,d_\beta}
\le
e_K+\frac{\varepsilon_{\mathrm{app}}}{1-\bar\gamma_\beta},
\]
which proves Theorem~\ref{thm:completeness-realizability}.
\end{proof}

\begin{proof}[Proof of Corollary~\ref{cor:linear-endpoint-refinement}]
Let
\[
e_k:=\|\widehat Q^{(k)}-Q^\pi_{\mathcal F,\beta}\|_{2,d_\beta}.
\]
Equation~\eqref{eq:bellman-residual-realizability},
Proposition~\ref{prop:linear-residual-interpolation}, and Cauchy--Schwarz give
\[
\varepsilon_{\mathrm{Bell},w}(\widehat Q^{(k)})
\le
\sqrt{\kappa_{\mathrm{res}}}\,\varepsilon_w
\min\left\{
\varepsilon_{\mathrm{Bell}},
\bar\gamma_\beta e_k+\frac{\varepsilon_{\mathrm{app}}}{1-\bar\gamma_\beta}
\right\}.
\]
Applying Lemma~\ref{lemma::errorperiter} at confidence level \(\tau/K\), taking
a union bound, and using Lemma~\ref{lem:bellman-contraction} yield, simultaneously
for \(0\le k<K\),
\begin{equation}
\label{eq:linear-endpoint-recursion}
e_{k+1}
\le
\bar\gamma_\beta e_k
+C\frac{\varepsilon_{\mathrm{stat}}(K,\tau)}
{\sqrt{\lambda_{\mathcal F}}}
+C\frac{\varepsilon_w}{\lambda_{\mathcal F}}
\min\left\{
\varepsilon_{\mathrm{Bell}},
\bar\gamma_\beta e_k+\frac{\varepsilon_{\mathrm{app}}}{1-\bar\gamma_\beta}
\right\}.
\end{equation}
Using the completeness branch of \eqref{eq:linear-endpoint-recursion} and
\(\bar\gamma_\beta\le\gamma_w\) gives
\[
e_{k+1}
\le
\gamma_w e_k
+C\frac{\varepsilon_{\mathrm{stat}}(K,\tau)}
{\sqrt{\lambda_{\mathcal F}}}
+C\frac{\varepsilon_w}{\lambda_{\mathcal F}}
\varepsilon_{\mathrm{Bell}}.
\]
Using the realizability branch and the definition of \(\gamma_w\) gives
\[
e_{k+1}
\le
\gamma_w e_k
+C\frac{\varepsilon_{\mathrm{stat}}(K,\tau)}
{\sqrt{\lambda_{\mathcal F}}}
+C\frac{\varepsilon_w}{\lambda_{\mathcal F}}
\frac{\varepsilon_{\mathrm{app}}}{1-\bar\gamma_\beta}.
\]
Taking the smaller bound and unrolling the resulting recursion gives
\[
e_K
\le
\gamma_w^K e_0
+\frac{C}{1-\gamma_w}
\left[
\frac{\varepsilon_{\mathrm{stat}}(K,\tau)}
{\sqrt{\lambda_{\mathcal F}}}
+\frac{\varepsilon_w}{\lambda_{\mathcal F}}
\min\left\{
\varepsilon_{\mathrm{Bell}},
\frac{\varepsilon_{\mathrm{app}}}{1-\bar\gamma_\beta}
\right\}
\right].
\]
Finally,
\(\|\widehat Q^{(K)}-Q^\pi\|_{2,d_\beta}
\le e_K+\varepsilon_{\mathrm{app}}/(1-\bar\gamma_\beta)\) by
Theorem~\ref{lem:proj-fixed-point-error}, proving
Corollary~\ref{cor:linear-endpoint-refinement}.
\end{proof}

\begin{proof}[Proof of Theorem~\ref{theorem:reward-perturbation}]
By Lemma~\ref{lemma::projected_contraction}, both projected Bellman operators
are \(\gamma\)-contractions on \(\mathcal F\); changing the reward does not
affect differences of Bellman updates. Their fixed-point identities and the
triangle inequality therefore give
\begin{equation}
\label{eq:reward-perturb-1}
(1-\gamma)\|\widehat Q^\pi_{\mathcal F}-Q^\pi_{\mathcal F}\|_{2,\mu_\pi}
\le
\|\widehat{\mathcal{T}}_{\mathcal F}(\widehat Q^\pi_{\mathcal F})
-\mathcal{T}_{\mathcal F}(\widehat Q^\pi_{\mathcal F})\|_{2,\mu_\pi}.
\end{equation}

Next, for any \(Q\),
\begin{align*}
\|\widehat{\mathcal{T}}_{\mathcal F}(Q) - \mathcal{T}_{\mathcal F}(Q)\|_{2,{\mu_\pi}}
&=
\|\Pi_{\mathcal F}\widehat{\mathcal{T}}(Q)
  - \Pi_{\mathcal F}\mathcal{T}(Q)\|_{2,{\mu_\pi}}
\\
&\le
\|\widehat{\mathcal{T}}(Q) - \mathcal{T}(Q)\|_{2,{\mu_\pi}}
\\
&=
\|\widehat r - r_0\|_{2,{\mu_\pi}},
\end{align*}
because
\(\widehat{\mathcal T}(Q)-\mathcal T(Q)=\widehat r-r_0\) and
\(\Pi_{\mathcal F}\) is nonexpansive in \(L^2(\mu_\pi)\).
Substituting into \eqref{eq:reward-perturb-1} yields
\[
\|\widehat Q^\pi_{\mathcal F} - Q^\pi_{\mathcal F}\|_{2,{\mu_\pi}}
\le
\frac{1}{1-\gamma}\,
\|\widehat r - r_0\|_{2,{\mu_\pi}}.
\]

\paragraph{Linear subspaces.}
If \(\mathcal F\) is a linear subspace of \(L^2({\mu_\pi})\), then
\(\Pi_{\mathcal F}\) is a linear orthogonal projection. Thus, for every
\(Q\in L^2(\mu_\pi)\),
\[
\widehat{\mathcal T}_{\mathcal F}(Q)-\mathcal T_{\mathcal F}(Q)
=
\Pi_{\mathcal F}(\widehat r-r_0).
\]
Substituting this into \eqref{eq:reward-perturb-1} yields
\[
\|\widehat Q^\pi_{\mathcal F} - Q^\pi_{\mathcal F}\|_{2,{\mu_\pi}}
\le
\frac{1}{1-\gamma}\,
\|\Pi_{\mathcal F}(\widehat r - r_0)\|_{2,{\mu_\pi}}.
\]
This proves the second bound.
\end{proof}

\section{FORE Guarantees for Occupancy-Ratio Estimation and FQE}
\label{app:fore-weight-bound}

This appendix has two goals. First, it states a finite-sample error bound for
the fitted occupancy-ratio evaluation (FORE) estimator. Second, it verifies
that the resulting weights satisfy the conditions of the FQE theorem. Combining
the two bounds proves Corollary~\ref{cor:fore-fqe-end-to-end}.

\subsection{FORE Estimator}
\label{app:fore-estimator}

Write \(X=(S,A)\). Each observed state--action pair generates two
target-policy draws. The first keeps the observed state, draws
\(A^0\sim\pi(\cdot\mid S)\), and sets \(X^0=(S,A^0)\). The second advances
the system one step:
\[
S'\sim P(\cdot\mid X),
\qquad
A'\sim\pi(\cdot\mid S').
\]
Let \(P_\pi(\cdot\mid x)\) denote the resulting target-policy kernel on
\(\mathcal S\times\mathcal A\). For a normalized ratio \(w\), define the
adjoint occupancy map
\[
\mathsf B_\beta^\pi w
:=
(1-\beta)\omega_{\pi/b}
+\beta\frac{\mathrm d\{(w\nu_b)P_\pi\}}{\mathrm d\nu_b}.
\]
It maps the current weighted distribution to one discounted occupancy update.
The target ratio is its fixed point:
\(\mathsf B_\beta^\pi w_\beta=w_\beta\).

For nonnegative functions \(f\) and \(g\), with \(g>0\), write
\[
D_{\nu_b}^{\rm gen}(f\|g)
:=
\mathbb E_{\nu_b}\!\left[
f(X)\log\frac{f(X)}{g(X)}-f(X)+g(X)
\right].
\]
Let \(\mathcal G_w\) be a class of log-ratio functions and define the
associated normalized exponential class
\[
\mathcal W_{\rm F}
:=
\left\{
w_g(x)
=
\frac{\exp\{g(x)\}}
{\mathbb E_{\nu_b}[\exp\{g(X)\}]}:
g\in\mathcal G_w
\right\}.
\]
When both arguments integrate to one, this is the ordinary KL divergence
between the corresponding distributions. In particular,
\[
D_{\nu_b}^{\rm gen}(w\|w_\beta)
=
D_{\rm KL}(w\nu_b\|d_\beta).
\]
Define the population approximation error
\[
\varepsilon_{\rm KL}
:=
\inf_{w\in\mathcal W_{\rm F}}
D_{\nu_b}^{\rm gen}(w\|w_\beta).
\]
Thus, \(\varepsilon_{\rm KL}=0\) whenever
\(\log w_\beta\) is represented by \(\mathcal G_w\) up to an additive
constant.

\begin{samepage}
Let
\[
\mathcal D_{n_w}^{w}
=
\{(X_i,X_i^0,X_i^+)\}_{i=1}^{n_w}
\]
denote the sample used to fit the weights, and let
\(\mathbb P_{n_w}^{w}\) denote its empirical average.
\end{samepage}
FORE
\citep{van2026fitted} initializes
\(\widehat w^{(0)}\equiv1\) and, for \(k=0,\ldots,K_w-1\), computes
\[
\begin{aligned}
\widehat g_{k+1}
&\in
\argmin_{g\in\mathcal G_w}
\left[
\log\mathbb P_{n_w}^{w}\!\left\{e^{g(X)}\right\}
-(1-\beta)\mathbb P_{n_w}^{w}\{g(X^0)\}
-\beta
\frac{
\mathbb P_{n_w}^{w}
\!\left\{\widehat w^{(k)}(X)g(X^+)\right\}
}{
\mathbb P_{n_w}^{w}
\!\left\{\widehat w^{(k)}(X)\right\}
}
\right],
\\
\widehat w^{(k+1)}(x)
&=
\frac{\exp\{\widehat g_{k+1}(x)\}}
{\mathbb P_{n_w}^{w}
 [\exp\{\widehat g_{k+1}(X)\}]}.
\end{aligned}
\]
The two linear terms target the discounted mixture of the restart sample
\(X^0\) and the weighted one-step sample \(X^+\). The denominator in the
second term normalizes the current weighted empirical distribution. This is
the discounted FORE recursion with source distribution \(\xi_{\pi,b}\).

\subsection{Conditions and Localized Complexity}
\label{app:fore-conditions}

\paragraph{Conditions.}
In addition to Conditions~\ref{cond::stationary} and
\ref{cond::fore-target-ratio-bounded}, FORE requires source coverage:
\(\xi_{\pi,b}\ll\nu_b\) and
\(\underline\omega_{\pi/b}>0\). The next two conditions specify the FORE
sample and function class. Condition~\ref{cond::split} is needed only when the
estimated weights are combined with FQE.
\begin{enumerate}[
label=\textbf{C\arabic*},
ref={C\arabic*},
leftmargin=1.5em,
resume=cond
]
\item \label{cond::fore-sample}
\textbf{FORE sample.}
The triples in \(\mathcal D_{n_w}^{w}\) are independent, with
\(X_i=(S_i,A_i)\sim\nu_b\). Conditionally on \(X_i\),
\(X_i^0=(S_i,A_i^0)\), where \(A_i^0\sim\pi(\cdot\mid S_i)\), and
\(X_i^+\sim P_\pi(\cdot\mid X_i)\), with the two draws independent.

\item \label{cond::fore-class}
\textbf{Bounded convex FORE class.}
The class \(\mathcal G_w\) is convex, compact in the supremum norm, contains
\(0\), and, for some \(R_w<\infty\), satisfies
\[
\sup_{g\in\mathcal G_w}
\left|
g(x)-\mathbb E_{\nu_b}\{g(X)\}
\right|
\le R_w
\qquad
\text{for \(\nu_b\)-almost every \(x\)}.
\]

\end{enumerate}

\paragraph{Interpretation.}
Convexity makes each update a convex fit on the log-ratio scale, while
compactness guarantees that the empirical minimum is attained. The centered
bound has the more transparent consequence
\[
e^{-2R_w}\le w(x)\le e^{2R_w}
\]
for every \(w\in\mathcal W_{\rm F}\) and every empirical FORE iterate. The
target-ratio bounds and occupancy balance equation also yield the one-step
coverage envelope
\begin{equation}
\label{eq:fore-automatic-smoothing}
\frac{d\{(w\nu_b)P_\pi\}}{d\nu_b}
\le
e^{2R_w}\frac{\overline\kappa_\beta}
{\beta\underline\kappa_\beta}
\qquad \nu_b\text{-almost surely}.
\end{equation}
These are the boundedness and one-step coverage quantities needed by the FORE
theorem. Their verification from the preceding conditions is included in the
proof of Theorem~\ref{thm:fore-stationary-weight}.

\paragraph{Localized critical radius.}
The FORE rate is summarized by one localized radius. It must control both the
log-ratio fit and the weighted one-step process created by the recursion. To
define it, introduce the centered and difference classes
\[
\mathcal G_w^\circ
:=
\left\{
g-\mathbb E_{\nu_b}g:
g\in\mathcal G_w
\right\},
\qquad
\mathcal G_w^\Delta
:=
\left\{
g_1-g_2:
g_1,g_2\in\mathcal G_w^\circ
\right\},
\]
and the induced multiplier class
\[
\mathcal M_w
:=
\left\{
(x,x^+)\mapsto w(x)g_\Delta(x^+):
w\in\mathcal W_{\rm F},\
g_\Delta\in\mathcal G_w^\Delta
\right\}.
\]
Let \(Q_{b,\Delta}\) denote the law of \((X,X)\) for
\(X\sim\nu_b\), and let \(Q_{b,\pi}\) denote the law of \((X,X^+)\) under
\[
X\sim\nu_b,
\qquad
X^+\mid X\sim P_\pi(\cdot\mid X).
\]

\begin{samepage}
For a square-integrable class \(\mathcal A\) under a distribution \(Q\),
define the localized Rademacher complexity
\[
\mathcal R_{n_w}(\mathcal A,r;Q)
:=
\mathbb E\!\left[
\sup_{\substack{a\in\mathcal A:\\
\|a\|_{L^2(Q)}\le r}}
\left|
\frac{1}{n_w}
\sum_{i=1}^{n_w}\sigma_i a(Z_i)
\right|
\right],
\]
where \(Z_i\stackrel{\rm iid}{\sim}Q\), and the
\(\sigma_i\)'s are independent Rademacher variables.
\end{samepage}
Set
\[
\begin{aligned}
\mathfrak C_{n_w}(r)
&:=
\max\Bigl\{
\mathcal R_{n_w}(\mathcal G_w^\Delta,r;\nu_b),
\mathcal R_{n_w}(\mathcal G_w^\Delta,r;\xi_{\pi,b}),
\mathcal R_{n_w}(\mathcal M_w,r;Q_{b,\Delta}),
\mathcal R_{n_w}(\mathcal M_w,r;Q_{b,\pi})
\Bigr\},
\\
\mathfrak r_{n_w,\rm fit}
&:=
n_w^{-1/2}
\vee
\inf\left\{
r>0:
\mathfrak C_{n_w}(r)\le r^2
\right\}.
\end{aligned}
\]

The radius \(\mathfrak r_{n_w,\rm fit}\) simultaneously controls the
log-ratio difference class and the two multiplier processes generated by the
FORE update.

\subsection{Weight-Error Bound for FORE}
\label{app:fore-weight-error-bound}

\begin{theorem}[Weight error of the FORE estimator]
\label{thm:fore-stationary-weight}
Assume Conditions~\ref{cond::stationary},
\ref{cond::fore-target-ratio-bounded}, and
\ref{cond::fore-sample}--\ref{cond::fore-class}. Also suppose that
\(\xi_{\pi,b}\ll\nu_b\) and \(\underline\omega_{\pi/b}>0\).
Let
\[
\widehat w_\beta:=\widehat w^{(K_w)},
\qquad
\varrho_\beta:=\frac{1+\beta}{2},
\]
and define
\[
\varepsilon_w
:=
\left\|
\frac{\widehat w_\beta}{w_\beta}-1
\right\|_{2,d_\beta}
=
\left\{
\mathbb E_{\nu_b}
\left[
\frac{(\widehat w_\beta-w_\beta)^2}{w_\beta}
\right]
\right\}^{1/2}.
\]
Set
\[
L_\beta
:=
1+\log\!\left(\frac{e^{2R_w}}{\underline\kappa_\beta}\right).
\]
There exists a finite constant
\(C_{{\rm FORE},\beta}^{\circ}
=C(R_w,\beta,\text{coverage constants})\) such that, for every
\(\tau\in(0,1)\), with probability at least \(1-\tau\),
\[
\begin{aligned}
D_{\nu_b}^{\rm gen}(\widehat w_\beta\|w_\beta)
\le C_{{\rm FORE},\beta}^{\circ}L_\beta\Bigg[
&\varrho_\beta^{K_w}
D_{\nu_b}^{\rm gen}(1\|w_\beta)
+
\frac{\varepsilon_{\rm KL}}
{(1-\beta)\underline\kappa_\beta}
\\
&+
\frac{\log^2(e n_w)}{(1-\beta)^2}
\left\{
\mathfrak r_{n_w,\rm fit}^2
+
\frac{\log(1/\tau)}{n_w}
\right\}
\Bigg].
\end{aligned}
\]
Consequently, after increasing this constant if necessary,
\[
\begin{aligned}
\varepsilon_w
\le C_{{\rm FORE},\beta}^{\circ}\Bigg[
&\frac{\varrho_\beta^{K_w/2}}
{\sqrt{\underline\kappa_\beta}}
\left\{
D_{\nu_b}^{\rm gen}(1\|w_\beta)
\right\}^{1/2}
+
\frac{1}{\underline\kappa_\beta}
\sqrt{\frac{\varepsilon_{\rm KL}}{1-\beta}}
\\
&+
\frac{\log(e n_w)}
{(1-\beta)\sqrt{\underline\kappa_\beta}}
\left\{
\mathfrak r_{n_w,\rm fit}
+
\sqrt{\frac{\log(1/\tau)}{n_w}}
\right\}
\Bigg].
\end{aligned}
\]
\end{theorem}

The generalized-KL inequality is the native FORE guarantee. The second
inequality converts it to the relative \(L^2(d_\beta)\) error used by the FQE
analysis. Its three terms are, respectively, the finite-iteration,
ratio-class approximation, and statistical errors.

\subsection{Proof of the End-to-End Guarantee for FQE with FORE Weights}
\label{app:fore-fqe-end-to-end}

\begin{proof}[Proof of Corollary~\ref{cor:fore-fqe-end-to-end}]
We first verify the two weight conditions required by
Theorem~\ref{thm:completeness-realizability}, then combine the FORE and FQE
probability bounds. Pointwise inequalities below are understood
\(\nu_b\)-almost surely. Set
\[
\overline\kappa_{\rm F}
:=
\max\{\overline\kappa_\beta,e^{2R_w}\},
\qquad
\lambda_{\mathcal F}
:=
\frac{e^{-2R_w}}{\overline\kappa_\beta}.
\]
\emph{Step 1: Uniform weight bounds.}
For any log-ratio fitted by FORE, define
\[
g^\circ
:=
g-\mathbb E_{\nu_b}g.
\]
Condition~\ref{cond::fore-class} gives
\[
-R_w\le g^\circ(x)\le R_w.
\]
Because empirical normalization is invariant to additive constants, every
fitted iterate can be written as
\[
\widehat w^{(k)}(x)
=
\frac{e^{g^\circ(x)}}
{\mathbb P_{n_w}^{w}e^{g^\circ(X)}}.
\]
Both the numerator and the empirical denominator lie in
\([e^{-R_w},e^{R_w}]\). Hence
\[
e^{-2R_w}
\le
\widehat w^{(k)}(x)
\le
e^{2R_w}.
\]
Condition~\ref{cond::fore-target-ratio-bounded} gives
\[
0<\underline\kappa_\beta\le w_\beta(x)\le\overline\kappa_\beta.
\]
Together, these bounds verify Condition~\ref{cond::weight-coverage} with
\(\overline\kappa_{\rm F}\).

\medskip
\noindent\emph{Step 2: Restricted norm comparability.}
For every \(h\in\mathcal F-\mathcal F\) satisfying
\(\|h\|_{2,d_\beta}>0\),
\[
\begin{aligned}
\frac{
\mathbb E_{\nu_b}[\widehat w_\beta h^2]
}{
\mathbb E_{\nu_b}[w_\beta h^2]
}
&\ge
\frac{
e^{-2R_w}\mathbb E_{\nu_b}[h^2]
}{
\overline\kappa_\beta\mathbb E_{\nu_b}[h^2]
}
\\
&=
\lambda_{\mathcal F}.
\end{aligned}
\]
Thus Condition~\ref{cond::weight-stability} holds with the stated
\(\lambda_{\mathcal F}\).

\medskip
\noindent\emph{Step 3: Combine the FORE and FQE bounds.}
Apply Theorem~\ref{thm:fore-stationary-weight} with failure probability
\(\tau/2\). On the resulting event,
\[
\varepsilon_w
\le
\varepsilon_{\rm FORE}(K_w,n_w,\tau/2).
\]
Conditional on the FORE sample, apply
Theorem~\ref{thm:completeness-realizability} with failure probability
\(\tau/2\).
In this application, Condition~\ref{cond::weight-coverage} holds with
\(\kappa_{\rm cov}=\overline\kappa_{\rm F}\), so
\(\varepsilon_{\mathrm{stat}}\) is evaluated with this envelope.
Both branches of the minimum in
Theorem~\ref{thm:completeness-realizability} are nondecreasing in
\(\varepsilon_w\). Substituting the FORE bound and the preceding specialization
gives the two replacements stated in
Corollary~\ref{cor:fore-fqe-end-to-end}.
Condition~\ref{cond::split} makes the FORE and FQE samples independent, so a
union bound gives joint probability at least \(1-\tau\), proving the claim.
\end{proof}

\subsection{Proof of the Weight-Error Bound for FORE}
\label{app:proof-fore-stationary-weight}
\enlargethispage{4\baselineskip}

\begin{proof}
The argument has two parts. We first specialize the discounted FORE theorem to
obtain its native generalized-KL bound. We then convert that bound to the
relative \(L^2(d_\beta)\) metric required by FQE, tracking the target-ratio
floor in both conversions.

\emph{Step 1: Generalized-KL error.}
The balance equation gives \(\mathsf B_\beta^\pi w_\beta=w_\beta\). For
normalized density ratios \(w\) and \(v\), the two images under
\(\mathsf B_\beta^\pi\) share the same restart component. Joint convexity of
KL and data processing under \(P_\pi\) therefore give
\[
D_{\nu_b}^{\rm gen}
\bigl(\mathsf B_\beta^\pi w\|\mathsf B_\beta^\pi v\bigr)
\le
\beta D_{\rm KL}\bigl((w\nu_b)P_\pi\|(v\nu_b)P_\pi\bigr)
\le
\beta
D_{\nu_b}^{\rm gen}(w\|v).
\]
Consequently,
\[
D_{\nu_b}^{\rm gen}
\bigl(\mathsf B_\beta^\pi w\|w_\beta\bigr)
\le
\beta D_{\nu_b}^{\rm gen}(w\|w_\beta).
\]

The cited FORE theorem measures approximation on the log-ratio scale, whereas
our statement uses generalized KL. To connect the two, fix
\(w\in\mathcal W_{\rm F}\) and set \(t=w/w_\beta\).
Conditions~\ref{cond::fore-class} and
\ref{cond::fore-target-ratio-bounded} imply
\[
\frac{e^{-2R_w}}{\overline\kappa_\beta}
\le t\le
\frac{e^{2R_w}}{\underline\kappa_\beta}.
\]
Let \(h(t):=t\log t-t+1\), with
\(h(t)/(\log t)^2:=1/2\) at \(t=1\). Since
\(h(t)/(\log t)^2\to\infty\) as \(t\to\infty\),
\[
c_{\rm KL}
:=
\inf_{t\ge e^{-2R_w}/\overline\kappa_\beta}
\frac{h(t)}{(\log t)^2}
>0.
\]
Thus \(c_{\rm KL}\) depends only on \(R_w\) and
\(\overline\kappa_\beta\), not on
\(\underline\kappa_\beta\), and
\(h(t)\ge c_{\rm KL}(\log t)^2\) throughout the displayed interval.
Therefore
\[
\|\log w-\log w_\beta\|_{2,\nu_b}^2
\le
\frac{1}{\underline\kappa_\beta c_{\rm KL}}
D_{\nu_b}^{\rm gen}(w\|w_\beta).
\]
Taking infima shows that the squared log-ratio approximation error in the
FORE theorem is at most
\(\varepsilon_{\rm KL}/
(\underline\kappa_\beta c_{\rm KL})\).

We next verify the remaining inputs to the cited result. By
Condition~\ref{cond::fore-sample}, the transition triples are independent.
The empirical objective is continuous in the supremum norm, so compactness in
Condition~\ref{cond::fore-class} guarantees exact minimizers and also gives the
required \(L^2(\nu_b)\) total boundedness. The same condition gives the uniform
ratio bound \(w\le e^{2R_w}\).

For completeness, the one-step envelope in
\eqref{eq:fore-automatic-smoothing} follows directly from the target-ratio
bounds and the occupancy balance equation. As inequalities between measures,
\[
\nu_b
\le \underline\kappa_\beta^{-1}d_\beta,
\qquad
d_\beta P_\pi
\le \beta^{-1}d_\beta
\le \beta^{-1}\overline\kappa_\beta\nu_b.
\]
Therefore, for \(w\in\mathcal W_{\rm F}\),
\[
(w\nu_b)P_\pi
\le
e^{2R_w}\frac{\overline\kappa_\beta}
{\beta\underline\kappa_\beta}\nu_b,
\]
which proves \eqref{eq:fore-automatic-smoothing}. Together with source
coverage, these facts verify the bounded-log-ratio, positivity, lower-tail,
and one-step empirical-process conditions of Theorem~4.2 in
\citet{van2026fitted}. Applying that theorem with source distribution
\(\xi_{\pi,b}\), discount \(\beta\), and the contraction above gives
\[
\begin{aligned}
D_{\nu_b}^{\rm gen}(\widehat w_\beta\|w_\beta)
\le C_{{\rm FORE},\beta}^{\circ}L_\beta\Bigg[
&\varrho_\beta^{K_w}
D_{\nu_b}^{\rm gen}(1\|w_\beta)
+
\frac{\varepsilon_{\rm KL}}
{(1-\beta)\underline\kappa_\beta}
\\
&+
\frac{\log^2(e n_w)}{(1-\beta)^2}
\left\{
\mathfrak r_{n_w,\rm fit}^2
+
\frac{\log(1/\tau)}{n_w}
\right\}
\Bigg],
\end{aligned}
\]
where
\[
\varrho_\beta=\frac{1+\beta}{2},
\qquad
L_\beta
=
1+\log\!\left(\frac{e^{2R_w}}{\underline\kappa_\beta}\right),
\]
after absorbing \(c_{\rm KL}^{-1}\) into the residual constant
\(C_{{\rm FORE},\beta}^{\circ}\).
Here the same sample may be reused across all \(K_w\) iterations because the
empirical-process bound is uniform over the log-ratio and induced multiplier
classes.

\medskip
\noindent\emph{Step 2: Relative \(L^2(d_\beta)\) error.}
It remains to convert generalized KL error to the relative error required by
the FQE theorem. Conditions~\ref{cond::fore-class} and
\ref{cond::fore-target-ratio-bounded} give
\[
\widehat w_\beta\le e^{2R_w},
\qquad
w_\beta\ge\underline\kappa_\beta.
\]
The one-sided generalized-KL-to-chi-square inequality of
\citet{van2026fitted}, applied with
\(a=\widehat w_\beta\), \(b=w_\beta\),
\(M=e^{2R_w}\), and \(m=\underline\kappa_\beta\), yields
\[
\varepsilon_w
\le
C_\chi
\left\{
D_{\nu_b}^{\rm gen}(\widehat w_\beta\|w_\beta)
\right\}^{1/2},
\]
where \(C_\chi<\infty\) depends only on
\(R_w\) and \(\underline\kappa_\beta\).

To expose the floor dependence in the theorem, combine this constant with
\(L_\beta\). Let \(U=e^{2R_w}/\underline\kappa_\beta\) and
\(\phi(t)=t\log t-t+1\). The proof of the cited inequality gives
\[
C_\chi
=
\left\{
\inf_{0\le t\le U}
\frac{\phi(t)}{(t-1)^2}
\right\}^{-1/2}.
\]
Here the ratio is defined by continuity at \(t=1\). It is nonincreasing, so
\(C_\chi=(U-1)/\sqrt{\phi(U)}\). Moreover,
\[
\sup_{U\ge1}
\frac{(U-1)\sqrt{1+\log U}}
{\sqrt{U\phi(U)}}
<\infty.
\]
A direct one-variable calculation therefore gives
\[
C_\chi\sqrt{L_\beta}
\le
\frac{C(R_w)}{\sqrt{\underline\kappa_\beta}}.
\]

Taking square roots in the preceding generalized-KL bound and using
subadditivity of the square root gives
\[
\begin{aligned}
\varepsilon_w
\le C_{{\rm FORE},\beta}^{\circ}\Bigg[
&\frac{\varrho_\beta^{K_w/2}}
{\sqrt{\underline\kappa_\beta}}
\left\{
D_{\nu_b}^{\rm gen}(1\|w_\beta)
\right\}^{1/2}
+
\frac{1}{\underline\kappa_\beta}
\sqrt{\frac{\varepsilon_{\rm KL}}{1-\beta}}
\\
&+
\frac{\log(e n_w)}
{(1-\beta)\sqrt{\underline\kappa_\beta}}
\left\{
\mathfrak r_{n_w,\rm fit}
+
\sqrt{\frac{\log(1/\tau)}{n_w}}
\right\}
\Bigg],
\end{aligned}
\]
after increasing the same constant if necessary.
Together with the generalized-KL bound, this proves both conclusions of
Theorem~\ref{thm:fore-stationary-weight}.
\end{proof}

\end{document}